\documentclass[10pt,twocolumn,prl,aps,floatfix,superscriptaddress,longbibliography]{revtex4-2}

\usepackage{xcolor}
\usepackage{hyperref}

\usepackage{graphicx}
\usepackage{physics}

\usepackage{amsthm}
\usepackage{amsmath}
\usepackage{amssymb}
\usepackage{calligra}

\usepackage{bm}
\usepackage{mathtools}
\usepackage[capitalise]{cleveref}
\usepackage{placeins}

\usepackage[T1]{fontenc}

\usepackage{calligra}
\DeclareMathAlphabet{\mathcalligra}{T1}{calligra}{m}{n}
\DeclareFontShape{T1}{calligra}{m}{n}{<->s*[2.2]callig15}{}

\DeclareSymbolFont{usualmathcal}{OMS}{cmsy}{m}{n}
\DeclareSymbolFontAlphabet{\mathcal}{usualmathcal}

\usepackage{tikz} %tikz is used to draw the glyphs
\usetikzlibrary{arrows,scopes} %tikz libraries used to draw the glyphs

\usepackage[color=orange!60,textsize=scriptsize]{todonotes}
\usepackage{siunitx}
\DeclareSIUnit\angstrom{\text{Å}}

\usepackage{pdfpages} % include pdfs
\usepackage{pgffor} % for loops
\usepackage{xr} % referencing the supplement

\makeatletter
\AtBeginDocument{\let\LS@rot\@undefined}
\makeatother

\def\supplementfilename{hessian_supplement}

\newif\ifarXiv
\arXivtrue 
\ifarXiv
    \pdfximage{\supplementfilename.pdf}
    \def\numbersupplementpages{\the\pdflastximagepages}
\fi

\begin{document}

\title{Explaining Near-Zero Hessian Eigenvalues Through Approximate Symmetries in Neural Networks}

\author{Marcel K\"uhn}
\affiliation{Institut f\"ur Theoretische Physik, Universit\"at Leipzig, 04103 Leipzig, Germany}
\affiliation{ScaDS.AI Dresden/Leipzig, Universit\"at Leipzig, 04105 Leipzig, Germany}

\author{Bernd Rosenow}
\affiliation{Institut f\"ur Theoretische Physik, Universit\"at Leipzig, 04103 Leipzig, Germany}

\begin{abstract}
The Hessian of the training loss governs the local geometry of the loss landscape, yet despite existing explanations for its largest eigenvalues, the origin of the vast multitude of vanishingly small eigenvalues remains elusive. We argue that the bulk consists of the weakly lifted pseudo-Goldstone modes of the continuous symmetries of the network parametrization. In deep linear networks these symmetries are exact: they generate flat directions and hence exact zero modes, whose eigenvectors we construct explicitly. Introducing a ReLU nonlinearity as a perturbation, we show that it breaks these symmetries weakly and explicitly. Resolving the spectrum at the level of eigenvectors, we find that the high-curvature directions are orthogonal to the symmetry subspace, while the bulk lies almost entirely within it. We demonstrate the mechanism in a two-layer ReLU student--teacher model and in a network trained on CIFAR-10. A convolutional example demonstrates that the same diagnostic extends beyond fully connected layers. Together, these results link the Hessian bulk to weakly broken symmetries and clarify the origin of near-zero modes.
\end{abstract}
\maketitle

\emph{Introduction---}The geometry of the loss landscape, encoded in the Hessian $\sf H$ of the training loss, underlies much of our understanding of how deep networks optimize and generalize~\cite{LeCun.1991, Geiger.2019, Becker.2020, Atanasov.2025, Cohen.2021}. It sets the local curvature relevant to second-order optimization~\cite{Amari.1998, Martens.2010, Martens.2015, Yao.2021}, is repeatedly tied to generalization through the flatness of minima~\cite{Hochreiter.1997, Keskar.2017, Foret.2021, Chaudhari.2019, Jastrzebski.2018}, and appears throughout pruning, robustness, and mode-connectivity analyses~\cite{Lecun.1989, Hassibi.1992, Singh.2020, Frantar.2023, Yao.2018, Moosavi.2019, Singla.2020, Garipov.2018, Brea.2019, Ainsworth.2023, Ito.2025}.

Empirically, the Hessian spectrum of a deep network is highly anisotropic: a dense bulk of small and nearly vanishing eigenvalues sits below a few large outliers~\cite{Sagun.2016, Sagun.2017, Ghorbani.2019}, the latter linked to data and class structure~\cite{Papyan.2019, Papyan.2020}. Analytic studies of shallow ReLU Hessians exploiting permutation symmetries~\cite{Arjevani.2020, Arjevani.2021} and random-matrix analyses of the closely related Fisher matrix~\cite{Pennington.2018, Karakida.2019, Karakida.2021} confirm that nonlinear networks generically carry many small but nonzero eigenvalues. What has been missing is a single, identifiable origin for this bulk.

We propose that, in fully connected networks in particular, the bulk is the spectral signature of weakly broken continuous symmetries of the specific architecture. Linear multilayer networks possess exact symmetries---inter-layer transformations that leave the function unchanged---which generate flat directions and thus exact zero modes. A nonlinearity like standard ReLU breaks these symmetries only weakly: the would-be flat directions acquire small curvature, exactly as an explicitly but weakly broken continuous symmetry produces parametrically light pseudo-Goldstone modes; in this analogy, the spontaneous symmetry breaking occurs already at network initialization.
Additional flat directions may arise from low input variance~\cite{LeCun.1991}, but often in far smaller number.
The picture connects naturally to the local geometry of loss landscapes~\cite{Fort.2019} and to gradient descent's confinement to a tiny top subspace~\cite{GurAri.2018}, since gradients are orthogonal to these symmetry directions.

Continuous and discrete network symmetries and their consequences for gradients and learning dynamics have been studied before~\cite{Du.2018, Brea.2019, Simsek.2021, Kunin.2021, Tanaka.2021, Zhao.2022, Marcotte.2023}, and degenerate Fisher matrices from parameter nonidentifiability are central to singular learning theory~\cite{Watanabe.2009, Fukumizu.2000}. In particular, the general identity between symmetries and Hessian eigenvectors~\cite{Kunin.2021} and the count of finite eigenvalues at the minimum of a deep linear network~\cite{Singh.2021, Bernacchia.2018} are established. In this Letter, we move from \emph{counting} to \emph{eigenvectors}, and from linear to nonlinear networks. We give an explicit orthogonal basis of symmetry generators in multilayer linear networks; we show, at the eigenvector level, that these symmetries persist approximately in nonlinear networks and after training; we establish an adiabatic connection between the nonlinear and linear spectra; and we extend the mechanism to a convolutional network.

% =====================================================================
\emph{Symmetries and zero eigenvalues---}We consider a model ${\bm f}$ mapping inputs ${\bf x}\in\mathbb{R}^N$ to outputs ${\bm f}({\bm \theta},{\bf x})\in\mathbb{R}^C$, with parameters ${\bm \theta}\in\mathbb{R}^d$ collecting all weights, and loss $\mathcal{L}({\bm \theta}) \coloneqq \langle \ell(\bm f({\bm \theta}, {\bf x}_i), {\bm y}_i) \rangle_i$ averaged over a dataset. The Hessian is ${\sf H} = \partial^2_{\bm \theta} \mathcal{L}$. A continuous symmetry is a transformation $\bm \phi(\alpha,{\bm \theta})$ with $\mathcal{L}({\bm \phi}(\alpha, {\bm \theta})) = \mathcal{L}({\bm \theta})$ and, for simplicity, ${\bm \phi}(0, {\bm \theta}) = {\bm \theta}$. Its generator is ${\bm \phi}' \coloneqq \left. \partial_\alpha {\bm \phi} \right|_{\alpha=0}$. We allow nonlinear symmetries (e.g. parameter products and inverses), as needed for the convolutional case below. For symmetries linear in the weights, ${\bm \phi}=\mathsf{R}(\alpha)\bm\theta$ with $\mathsf{R}(\delta\alpha)\approx\mathsf 1+\delta\alpha\,\mathsf G$, one has ${\bm \phi}'=\mathsf G\bm\theta$.

Differentiating the invariance condition $\left.\partial_\alpha \mathcal{L} \right|_{\alpha=0}=0$ with respect to ${\bm \theta}$ gives~\cite{Kunin.2021}
\begin{align}
    0 = {\sf H} {\bm \phi}' + \left[ \partial_{\bm \theta} ({\bm \phi}')^\top \right] \partial_{\bm \theta} \mathcal{L} \ .
    \label{eq:zero_eigenvector_general}
\end{align}
At critical points the second term vanishes and ${\bm \phi}'$ is a zero-eigenvalue eigenvector of $\sf H$: every continuous symmetry of the loss is a flat direction. Away from critical points, curvature along ${\bm \phi}'$ may be finite (see the rotation-invariant Mexican-hat potential, End Matter, \cref{fig:mexican_hat}).

% =====================================================================
\emph{Hessian, Gauss--Newton, and Fisher---}The architectural symmetries we construct are stronger than generic loss invariances: they leave the function ${\bm f}$ itself unchanged, so the network gradient is orthogonal to the symmetry generator, $\partial_{\bm \theta^{\!\top}}\! {\bm f}\, {\bm \phi}' = 0$, at \emph{every} ${\bm \theta}$. This matters because of the Gauss--Newton decomposition,
\begin{align}
    {\sf H} = \big\langle \partial_{\bm \theta^{\vphantom{\top}}} {\bm f}^{\!\top}_i \!\! \left(\partial_{\!\bm f}^2 \ell_i \right) \partial_{\bm \theta^{\!\top}}\! {\bm f}_i \big\rangle_i + \big\langle\! \left( \partial_{\!\bm f^{\! \top}} \ell_i \right)\partial^2_{\bm \theta}{\bm f}_i \big\rangle_i \ ,
    \label{eq:gauss_newton_decomposition}
\end{align}
with ${\bm f}_i \coloneqq {\bm f}({\bm \theta}, {\bf x}_i)$ and $\ell_i \coloneqq \ell({\bm f}_i, {\bm y}_i)$. The second term is a residual that vanishes when the outputs interpolate the labels and is small near critical points~\cite{Sagun.2017}. The first term, the generalized Gauss--Newton matrix ${\sf F}$, is positive semidefinite for losses convex in the output and obeys ${\sf F}\,{\bm \phi}' = 0$ at \emph{arbitrary} parameters, whereas \cref{eq:zero_eigenvector_general} guarantees Hessian zero modes only at critical points. For mean-squared error and softmax cross-entropy, ${\sf F}$ is the label-independent Fisher information matrix~\cite{Bishop.2006, Heskes.2000} and equals the Hessian of a self-target loss whose labels are the network's own outputs,
\begin{align}
    {\sf F} = \partial^2_{\bm\theta'} \big\langle \ell\big({\bm f}_i({\bm \theta}'),  {\bm f}_i({\bm \theta})\big)\big\rangle_i\Big|_{{\bm \theta}'={\bm \theta}} \ .
    \label{eq:fisher_self_target}
\end{align}
${\sf F}$ is in its own right the metric of natural-gradient optimization and a standard curvature probe~\cite{Amari.1998, Martens.2020, Pennington.2018}. The Hessian inherits these zero modes whenever the residual in \cref{eq:gauss_newton_decomposition} vanishes, as in the interpolating student--teacher experiments below.

\begin{figure}[t]
\centering
\includegraphics[width=0.98\columnwidth, trim=5 7 4 4, clip]{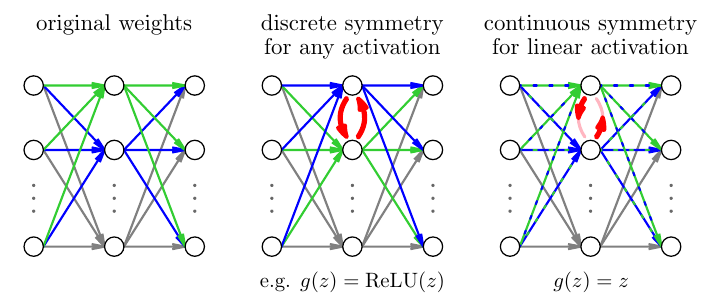}
\caption{Symmetries in fully connected layers. Left: original network, two hidden neurons highlighted. Middle: discrete permutation symmetry, swapping incoming and outgoing connections of two hidden units. Holds for any activation. Right: continuous symmetry of linear networks, an infinitesimal rotation of the incoming weights compensated on the outgoing weights. Preserves the output and generates a zero mode.}
\label{fig:mlp_symmetry}
\end{figure}

% =====================================================================
\emph{Linear networks: exact symmetries and their eigenvectors---}Consider the two-layer linear network ${\bm f}_{\rm lin}(\boldsymbol{\theta}, {\bf x}) = {\sf W}^{(2)} {\sf W}^{(1)} {\bf x}$, with ${\sf W}^{(1)} \in \mathbb{R}^{N \times N}$, ${\sf W}^{(2)} \in \mathbb{R}^{C \times N}$, and $\boldsymbol{\theta}$ stacking both ($d=N^2+NC$). Inserting any invertible ${\sf M}$ and its inverse between the layers leaves the output unchanged, so the network has a symmetry under the general linear group ${\rm GL}_N(\mathbb{R})$. We restrict to the positive-determinant subgroup, ${\rm GL}_N^+(\mathbb{R})$, containing the identity~\cite{Du.2018, Kunin.2021}. Its generators are the matrices ${\sf A}\in\mathbb{R}^{N\times N}$, acting through ${\sf M} = {\rm e}^{\alpha {\sf A}}$ as
\begin{align}
    {\sf W}^{(2)} {\rm e}^{-\alpha {\sf A}}{\rm e}^{\alpha {\sf A}}{\sf W}^{(1)} {\bf x} = {\sf W}^{(2)} {\sf W}^{(1)} {\bf x} \ ,
    \label{eq:layer_symmetry_exponential}
\end{align}
which maps the weights to ${\rm e}^{\alpha {\sf A}}{\sf W}^{(1)}$, ${\sf W}^{(2)} {\rm e}^{-\alpha {\sf A}}$ and gives the generators
\begin{align}
    \boldsymbol{\phi}_{\sf A}' = \begin{pmatrix} {\rm vec}\!\left({\sf A}{\sf W}^{(1)}\right) \\ {\rm vec}\!\left(-{\sf W}^{(2)}{\sf A} \right) \end{pmatrix} \ .
    \label{eq:lin_generators}
\end{align}
The $N^2$ independent ${\sf A}$ can be organized into canceling rescalings (diagonal ${\sf A}$), rotations (antisymmetric ${\sf A}$), and hyperbolic rotations (symmetric off-diagonal ${\sf A}$). The discrete permutation symmetry present even with nonlinear activations (\cref{fig:mlp_symmetry}) is not continuous and produces no zero eigenvalues.

Going beyond the count, we exhibit the eigenvectors explicitly. With the singular value decompositions ${\sf W}^{(k)} = {\sf U}^{(k)} {\sf S}^{(k)} ({\sf V}^{(k)})^{\!\top}$, $k=1,2$, the choice ${\sf A}^{(n,m)} \coloneqq {\bf v}_n^{(2)} ({\bf u}_m^{(1)})^{\!\top}$, consisting of the corresponding singular vectors, renders the generators \eqref{eq:lin_generators} mutually orthogonal (End Matter). The $N^2$ orthogonal generators span exactly the null space of $\sf H$. The symmetry thus induces precisely $N^2$ zero modes, which carry over to ${\sf F}$ at any parameters and any input distribution. The remaining $NC$ eigenvalues are generically finite, matching both the overparametrization (the input--output map is a single $C\times N$ matrix) and the known finite-eigenvalue count at the minimum of deep linear networks~\cite{Singh.2021, Bernacchia.2018}. 
Biases, absorbed by appending a constant input component, merely render this map affine, so $NC+C$ eigenvalues stay finite, with details in the Supplemental Material (SM)~\cite{supp}.
The construction extends to arbitrary layer sizes and additional layers. In particular, for hidden dimension $M$ with $N>M>C$, the two-layer network has $M^2+(N-M)(M-C)$ architectural invariances and still $NC$ finite modes, the additional $(N-M)(M-C)$ zero modes correspond to directions of ${\sf W}^{(1)}$ that map onto the kernel of the downstream map  ${\sf W}^{(2)}$~\cite{supp}.

%The construction extends to arbitrary layer sizes, additional layers, and biases (Supplement).

% =====================================================================
\emph{Weak breaking: pseudo-Goldstone modes---}While multilayer linear networks offer valuable theoretical insights \cite{Kawaguchi.2016, Singh.2021, Bernacchia.2018}, they are of limited practical relevance. 
To reach realistic networks while keeping analytic control, we switch on a nonlinearity as a perturbation, defining
\begin{align}
    \bm f_{\rm LReLU}^{(\varepsilon)}(\boldsymbol{\theta}, {\bf x}) \coloneqq {\sf W}^{(2)}\, g^{(\varepsilon)}\!\left({\sf W}^{(1)} {\bf x}\right) \ ,
\end{align}
with the Leaky-ReLU activation~\cite{Maas.2013} $g^{(\varepsilon)}(z) = z\,[1 - \varepsilon\, \Theta(-z)]$, where $\Theta$ is the Heaviside step. At $\varepsilon=0$ the network is linear, at $\varepsilon=1$ it is the standard ReLU. For positively homogeneous activations like Leaky-ReLU, the diagonal rescaling subgroup ($N$ generators) survives exactly for all $\varepsilon$~\cite{Kunin.2021}. The remaining $N^2-N$ symmetry directions are the ones the nonlinearity can break.

We study the student--teacher loss
\begin{align}
    \mathcal{L}^{(\varepsilon)}(\boldsymbol{\theta}) = \frac{1}{2C} \left\langle \left\lVert \bm f_{\rm LReLU}^{(\varepsilon)}(\boldsymbol{\theta}, {\bf x}) - \bm \tau^{(\varepsilon)}({\bf x}) \right\rVert^2\right\rangle_{\bf x} \ ,
    \label{eq:student_teacher_loss_lrelu}
\end{align}
over Gaussian inputs ${\bf x} \sim \mathcal{N}(0, {\sf 1}_N)$ with teacher $\bm \tau^{(\varepsilon)} \coloneqq \bm f_{\rm LReLU}^{(\varepsilon)}(\boldsymbol{\theta}_0, \cdot)$, and evaluate the Hessian at the global minimum $\boldsymbol{\theta}=\boldsymbol{\theta}_0$, where ${\sf H}^{(\varepsilon)}={\sf F}^{(\varepsilon)}$. The family interpolates continuously between ${\sf H}^{(0)}={\sf H}_{\rm lin}$ with exact symmetries and the fully nonlinear ${\sf H}^{(1)}={\sf H}_{\rm ReLU}$.

\begin{figure}[t]
\centering
\includegraphics{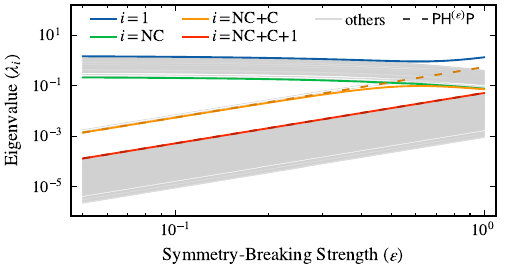}
\caption{Ranked Hessian eigenvalues of the two-layer Leaky-ReLU student--teacher model ($N=20$, $C=3$, single draw of Gaussian teacher weights ${\sf W}^{(k)}_{ij} \sim \mathcal{N}(0,\sigma^2)$ with $\sigma^2=1/N$) as a function of the symmetry-breaking strength $\varepsilon$, from the linear network ($\varepsilon\to0$) to ReLU ($\varepsilon=1$). ``$NC{+}C$'' and ``$NC{+}C{+}1$'' are the modes bordering the gap. The bulk eigenvalues, zero at $\varepsilon=0$, rise as $\varepsilon^2$ and follow the eigenvalues of the projected block ${\sf P}{\sf H}^{(\varepsilon)}{\sf P}$ (dashed lines). The $N$ rescaling zero modes stay at zero for all $\varepsilon$, below the plotted range. A small group of $C$ modes (orange, the mean-gradient outliers) bends near $\varepsilon\approx0.5$ to merge with the large block. The clean $\varepsilon^2$ trend without $\varepsilon^4$ bending indicates eigenvectors that stay within the symmetry subspace.}
\label{fig:epsilon_scaling}
\end{figure}

\emph{The lifted eigenvalues scale as $\varepsilon^2$---}\cref{fig:epsilon_scaling} tracks the full spectrum as $\varepsilon$ grows from $0$ to $1$. The eigenvalues that are exactly zero at $\varepsilon=0$ rise smoothly and---over almost the entire bulk---track $\varepsilon^2$ all the way to the ReLU point. This is sharper than it may look: since ${\sf H}^{(\varepsilon)}={\sf H}_{\rm lin}+\mathcal{O}(\varepsilon)$, one would naively expect a \emph{linear} rise, but the projection onto the symmetry subspace cancels the linear term, leaving $\varepsilon^2$ (End Matter). The $N$ rescaling modes remain pinned at zero for all $\varepsilon$. The  lifted eigenvalues thus behave as pseudo-Goldstone modes: would-be flat directions of a continuous symmetry, that is explicitly broken.

\emph{Bulk eigenvectors stay in the symmetry subspace---}An $\varepsilon^2$ law has a direct eigenvector consequence, which we both derive and measure. Splitting Leaky-ReLU into a linear and an absolute-value part, the Hessian of this two-layer model separates exactly, ${\sf H}^{(\varepsilon)} = (1-\tfrac\varepsilon2)^2{\sf H}_{\rm lin} + (\tfrac\varepsilon2)^2{\sf H}_{|\cdot|}$, the cross term vanishing by parity of the Gaussian input average (End Matter).
Beyond two layers and sign-symmetric inputs, the coupling between linear symmetry subspace and curvature subspace is generically $\mathcal{O}(\varepsilon)$, and only the $\varepsilon^2$ eigenvalue scaling survives~\cite{supp}. Here, however, the inter-space coupling is $\mathcal{O}(\varepsilon^2)$, so a bulk eigenvalue receives a mixing correction only at $\mathcal{O}(\varepsilon^4)$, with a coefficient that is a nonnegative quadratic form in the leakage of the mode into the curvature subspace. A mode that tracks $\varepsilon^2$ with \emph{no} $\varepsilon^4$ correction therefore has zero leakage: its eigenvector is confined to the symmetry subspace. The clean $\varepsilon^2$ scaling in \cref{fig:epsilon_scaling} is thus spectral evidence for confinement and yields an adiabatic connection between the ReLU Hessian at $\varepsilon=1$ and the linear one at $\varepsilon=0$.

\begin{figure}[t]
\centering
\includegraphics[trim=0 3 0 0, clip]{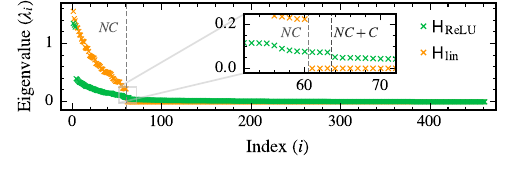}
\includegraphics[trim=0 6 0 0, clip]{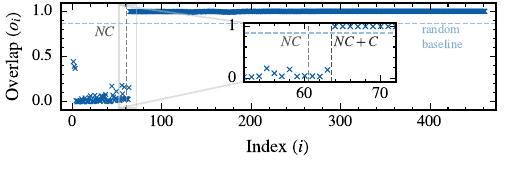}
\caption{Two-layer student--teacher model of \cref{fig:epsilon_scaling}, Hessian at the global minimum. Top: ReLU (green) and linear (orange) spectra. The linear network has $NC=60$ finite eigenvalues, the ReLU network 
has $C=3$ additional large eigenvalues due to nonzero $\langle \partial_{\bm\theta}{\bm f}^\top \rangle_{\bf x}$. Bottom: overlap $o_i$ \eqref{eq:overlap_definition} between ReLU eigenvectors and the linear symmetry subspace. Small-eigenvalue modes have $o_i\approx1$, revealing alignment with linear symmetry directions. The leading $NC+C$ modes are strongly suppressed (random baseline $o_{\rm random}\approx0.87$).}
\label{fig:shallow_relu_model}
\end{figure}

We confirm this directly at the ReLU point. For each eigenvector ${\bf p}_i$ of ${\sf H}_{\rm ReLU}$ we measure its overlap with the symmetry subspace,
\begin{align}
    o_i \coloneqq \lVert {\sf P}\,{\bf p}_i\rVert^2 \ 
    = \sum_{j\, :\, \text{EV}=0} \!\left({\bf p}_i \cdot {\bf p}_j^{(\rm lin)}\right)^{\!2} ,
    \label{eq:overlap_definition}
\end{align}
with ${\sf P}$ the projector onto the symmetry subspace. $o_i=1$ means ${\bf p}_i$ lies entirely within it. Because the subspace can be large, the random-vector baseline $o_{\rm random}=\mathrm{rank}({\sf P})/d$ with $\mathrm{rank}({\sf P})=N^2$ is high, and the informative signal is not $o_i\simeq1$ in the tail but the \emph{suppression} of $o_i$ for the high-curvature modes. \cref{fig:shallow_relu_model} shows the resulting two-tier structure: the ReLU spectrum has $NC+C\approx NC$ large eigenvalues separated by a small gap from a bulk of small ones, and the overlap is near zero for the $NC$ leading modes and near unity for the bulk. The symmetry directions of the linear network persist, approximately, as the bulk of the ReLU spectrum. An analogous three-layer student--teacher analysis in the SM~\cite{supp} reproduces this two-tier separation: the high-curvature modes remain clearly suppressed in the overlap, albeit less completely than in the shallow case, and the $\varepsilon^2$ scaling holds only approximately toward the ReLU point.

\emph{Additional outliers---}The $C$ extra large eigenvalues of the ReLU Hessian---the modes peeling off in \cref{fig:epsilon_scaling}---are the known mean-gradient outliers~\cite{Karakida.2019b, Papyan.2019}. ReLU units have positive average activations, so the network gradient has a finite mean $\langle \partial_{\bm\theta}{\bm f}^\top \rangle_{\bf x}$. Since ${\sf H}^{(\varepsilon)}={\sf F}^{(\varepsilon)}$ is the input-averaged gradient outer product, this mean contributes a rank-$C$ term producing the largest eigenvalues. In a tractable large-$N$ setting the leading ReLU eigenvector aligns with the mean gradient while retaining a substantial component along the linear symmetry directions~\cite{supp}.

% =====================================================================
\emph{A trained network on CIFAR-10---}We now test the mechanism away from the idealized minimum, on a network trained on real data. A three-layer fully connected ReLU network---$3072$-dimensional input, two hidden layers of $128$ units, $10$-way softmax, no biases, $410{,}880$ parameters---reaches $53\%$ test accuracy on CIFAR-10~\cite{Krizhevsky.2009}, as expected for a plain MLP~\cite{Zhang.2021}. We compute the Hessian of the cross-entropy training loss at the endpoint of training~\cite{supp}. While the endpoint is not an exact critical point, we still compare the training loss Hessian eigenvectors with the zero modes of a linear comparison.

For the linear comparison model we copy the trained ReLU parameters and remove the hidden nonlinearities but keep the softmax, whose invariance under uniform shifts of the $C$ preactivations~\cite{Kunin.2021} gives an effective output dimension $C_{\rm eff}\coloneqq C-1=9$ (End Matter). The input data carries a second, data-dependent source of flatness: low-variance input directions produce small Hessian eigenvalues~\cite{LeCun.1991}. To control this, we whiten the inputs~\cite{supp}, rescaling the top $100$ data covariance directions ($90\%$ of the cumulative variance) to unity and leaving the rest at much smaller variance~\cite{Bell.1997, Lee.2020}, which yields an effective input dimension $N_{\rm eff}=100$. The comparison subspace, collectively termed the symmetry subspace ${\sf P}$, is the null space of the linear model's Fisher with the low-variance directions set to zero variance. It combines the architectural symmetry directions with the suppressed-variance directions. Dimension counting makes this composition explicit: of the $d-N_{\rm eff}C_{\rm eff}=409{,}980$ directions in ${\sf P}$, the $d-NC_{\rm eff}=383{,}232$ architectural symmetries leave the model function unchanged for \emph{arbitrary} inputs.
The much smaller subset of $(N-N_{\rm eff})C_{\rm eff}=26{,}748$ zero modes alters the function only along suppressed-variance directions.

\begin{figure}[t]
\centering
\includegraphics[trim=0 6 0 0, clip]{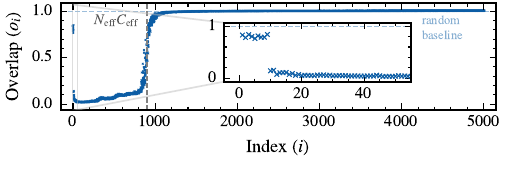}
\caption{Three-layer ReLU network trained on CIFAR-10, with effective dimensions $N_{\rm eff}=100$, $C_{\rm eff}=9$. Overlap $o_i$ \eqref{eq:overlap_definition} between the largest $5{,}000$ ReLU Hessian eigenvectors and the symmetry subspace ${\sf P}$ of the linear comparison model. Apart from $C_{\rm eff}=9$ leading outliers, only modes beyond the $N_{\rm eff}C_{\rm eff}=900$ largest reach $o_i\approx1$. The random baseline is $o_{\rm random}\approx0.998$ (standard deviation $\approx10^{-4}$), so the leading modes are suppressed by hundreds to thousands of standard deviations. The cumulative coverage \eqref{eq:coverage_definition} reaches $O_{5000}\approx0.95$, implying the unmeasured tail is almost entirely symmetric.}
\label{fig:mlp_cifar10_overlap}
\end{figure}

Combining effective dimensions, the linear model predicts only $N_{\rm eff}C_{\rm eff}=900$ high-curvature directions, i.e.\ $\mathrm{rank}({\sf Q})=900$ with ${\sf Q}={\sf 1}-{\sf P}$. \cref{fig:mlp_cifar10_overlap} shows the same two-tier structure as the student--teacher model, where the leading $N_{\rm eff}C_{\rm eff}=900$ ReLU modes have vanishing overlap with ${\sf P}$, the smaller eigenvalues have overlap near unity, and a set of $C_{\rm eff}=9$ outliers again shows raised overlap (mean-gradient mechanism). Here the suppression is dramatic: the random baseline is $o_{\rm random}\approx0.998$ with standard deviation $\approx10^{-4}$, so the leading modes are suppressed by thousands of standard deviations. With original data variances retained, the sharp transition is smoothed by covariance-induced eigenvalue spread, and the symmetry mechanism entangles with data-covariance flatness, although the leading high-curvature modes still show substantially reduced overlap~\cite{supp}.

Numerical constraints restrict us to the leading $5{,}000$ eigenvectors of the $410{,}880$-dimensional Hessian, seemingly leaving the overlap of the remaining $\sim\!4\times10^{5}$ modes unmeasured. We close this gap without diagonalizing them by defining the \emph{cumulative curvature coverage}
\begin{align}
    O_k \;\coloneqq\; \frac{1}{\mathrm{rank}({\sf Q})}\sum_{i=1}^{k}\,(1-o_i)
    \;=\; \frac{\mathrm{Tr}\!\left({\sf Q}\,{\sf \Pi}_k\right)}{\mathrm{Tr}({\sf Q})}\ \le\ 1 \ ,
    \label{eq:coverage_definition}
\end{align}
where ${\sf \Pi}_k=\sum_{i=1}^k{\bf p}_i{\bf p}_i^\top$ projects onto the top-$k$ eigenspace. $O_k$ is the fraction of the entire high-curvature subspace already spanned by the leading $k$ Hessian eigenvectors. We find $O_{5000}\approx0.95$: the first $5{,}000$ modes account for $95\%$ of the $900$ finite curvature directions of the linear network, so the remaining $\sim\!4\times10^{5}$ modes share only the leftover $5\%$ and must, on average, have overlap with the symmetry subspace very close to unity. Up to this $5\%$ residual, the bulk is symmetry-organized in its entirety.

% =====================================================================
\emph{A convolutional network---}Convolutional networks are a standard architecture for extracting local features, most notably in image processing \cite{Fukushima.1980, Lecun.1998, Goodfellow.2016}. To demonstrate that the same diagnostic applies beyond fully connected layers and beyond symmetries linear in the weights, we consider a one-dimensional convolutional network
\begin{align}
    f_{\rm conv}(\boldsymbol{\theta}, {\bf x}) = \big({\bf w}^{(2)}\big){\!}^\top\!\left({\bf w}^{(1)} * {\bf x}\right) \ ,
    \label{eq:convolution_network}
\end{align}
with kernel ${\bf w}^{(1)}\in\mathbb{R}^K$ and readout ${\bf w}^{(2)}\in\mathbb{R}^N$. With periodic boundary conditions the convolution is a circulant matrix ${\sf C}({\bf w}^{(1)})$, and inserting canceling exponentials as in \cref{eq:layer_symmetry_exponential} yields symmetries. The generator must now map the circulant to another circulant, which because ${\bf w}^{(1)}$ has only $K$ entries leaves exactly $K$ continuous symmetry directions. These symmetries are \emph{nonlinear} in the parameters (they involve the kernel through ${\sf C}^{-1}$), which is precisely why we allowed nonlinear $\bm\phi$ from the outset. The explicit construction and its two-dimensional extension are in the SM~\cite{supp}.

\begin{figure}[t]
\centering
\includegraphics[trim=0 6 0 0, clip]{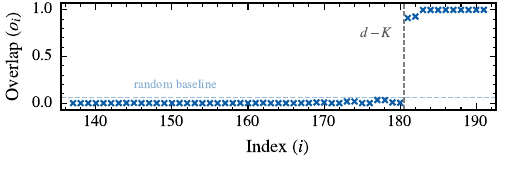}
\caption{Convolutional student--teacher network \eqref{eq:convolution_network}, input width $200$, hidden width $N=190$, kernel width $K=11$, mean-squared error, Hessian at identical student/teacher parameters drawn once from a Gaussian. Overlap between the eigenvectors of the nonperiodic-network Hessian and the $K=11$ symmetry directions of the corresponding periodic network, for the $60$ smallest eigenvalues. The random baseline is $K/d\approx0.055$ (dashed). Although the boundary conditions break the exact symmetry, the smallest $K$ modes retain near-unity overlap.}
\label{fig:conv_eigenspace}
\end{figure}

Here the symmetry is broken not by a nonlinear activation but by a different mechanism: nonperiodic boundary conditions, which enlarge the input width to $N+K-1$ and thereby reduce the overparametrization. This makes the convolutional model a test of the generality of the diagnostic rather than of the specific nonlinear mechanism. The conclusion is the same: in a nonperiodic student--teacher setup the $K$ smallest-eigenvalue eigenvectors retain near-unity overlap with the symmetry subspace of the periodic network (\cref{fig:conv_eigenspace}). Broken symmetries again leave identifiable low-curvature directions.

\emph{Conclusion---}We have proposed a mechanism for the Hessian bulk. The near-zero modes are the weakly lifted pseudo-Goldstone modes of architectural continuous symmetries: exact in linear networks, with explicitly constructed eigenvectors; weakly and explicitly broken by nonlinearity, with eigenvectors that remain largely confined to the symmetry subspace, adiabatically connected to the linear limit. Because the symmetry zero modes survive in the Fisher at arbitrary parameters, the same directions organize the Fisher and Gauss--Newton spectra, with consequences for natural-gradient and second-order methods~\cite{Amari.1998, Pascanu.2013, Martens.2020, Bernacchia.2018}.
The mechanism is expected to generalize beyond fully connected architectures to any model containing such layers, including the fully connected blocks of modern transformers. 
Altogether, these results provide theoretical grounding and practical intuition for interpreting the Hessian spectrum and optimization dynamics.

%%%%%%%%%%%%%%%%%%%%%%%%%%%%%%%%%%%%%%%%%%%%%%%%

\emph{Acknowledgments---}M.K. acknowledges financial support by the Federal Ministry of Research, Technology and Space of Germany and by the Sächsische Staatsministerium für Wissenschaft, Kultur und Tourismus in the programme Center of Excellence for AI-research ``Center for Scalable Data Analytics and Artificial Intelligence Dresden/Leipzig'', project identification number: ScaDS.AI.
The work of B.R.~is supported in part by the Simons Foundation Grant No.~12574. 

\emph{Data availability---}The data that support the findings of this article are openly available \cite{paperrepo}.

\bibliography{hessian_symmetry}

\onecolumngrid

\section{End Matter}

\twocolumngrid

\begin{figure}[t]
    \centering
    \includegraphics[width=0.47\columnwidth, trim=40 25 27 60, clip]{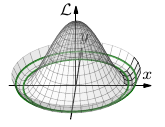}
    \includegraphics[width=0.47\columnwidth, trim=37 23 20 40, clip]{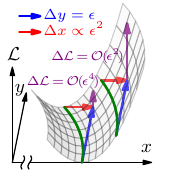}
    \caption{Zero Hessian eigenvalue from rotational symmetry, $\mathcal{L}(x,y) = (x^2 + y^2 - 1)^2$. Left: equipotential lines (green). Right: zoom on the minimum. At $\theta=(1,0)$ the symmetry direction ${\bf p}=(0,1)$ 
    gives rise to $\Delta\mathcal{L}=\mathcal{O}(\epsilon^4)$ with vanishing second derivative for $\varepsilon \to 0$. At $(x,0)$ with $x>1$, a shift $\Delta y=\epsilon$ (blue) is a rotation (green) plus a radial shift $\Delta x\propto\epsilon^2$ (red), raising the loss by $\mathcal{O}(\epsilon^2)$ with nonzero second derivative (finite Hessian eigenvalue).}
    \label{fig:mexican_hat}
\end{figure}

\emph{Appendix A: Curvature despite symmetry in the Mexican-hat picture---}The mechanism of \cref{eq:zero_eigenvector_general} is seen most simply in the rotation-invariant potential $\mathcal{L}(x,y) = (x^2 + y^2 - 1)^2$ (\cref{fig:mexican_hat}). At the minimum $(1,0)$ the symmetry direction $(0,1)$ is a flat zero mode. Away from the minimum, at $(x,0)$ with $x>1$, an infinitesimal move along $y$ is a rotation combined with an $\mathcal{O}(\epsilon^2)$ radial shift. Because the radial gradient is nonzero there, the loss increases at $\mathcal{O}(\epsilon^2)$ and the curvature along the symmetry direction becomes finite. Hence, symmetry directions are exactly flat at critical points but may acquire curvature off them, controlled by the residual term in \cref{eq:zero_eigenvector_general}.

\setcounter{equation}{0}
\renewcommand{\theequation}{B\arabic{equation}}

\emph{Appendix B: Orthogonal generator basis---}For the two-layer linear network, the choice ${\sf A}^{(n,m)} \coloneqq {\bf v}_n^{(2)} ({\bf u}_m^{(1)})^{\!\top}$ in \cref{eq:lin_generators}, with the singular value decompositions ${\sf W}^{(1)} = {\sf U}^{(1)} {\sf S}^{(1)} ({\sf V}^{(1)})^{\!\top}$ and ${\sf W}^{(2)} = {\sf U}^{(2)} {\sf S}^{(2)} ({\sf V}^{(2)})^{\!\top}$, gives
\begin{subequations}
\label{eq:lin_generators_svd}
\begin{alignat}{2}
    {\sf A}^{(n,m)} {\sf W}^{(1)} &= s_m^{(1)}\, {\bf v}_n^{(2)} \big({\bf v}_m^{(1)}\big)^{\!\top}, \\
    {\sf W}^{(2)} {\sf A}^{(n,m)} &= s_n^{(2)}\, {\bf u}_n^{(2)} \big({\bf u}_m^{(1)}\big)^{\!\top},
\end{alignat}
\end{subequations}
where $s_m^{(i)}$ is a singular value, ${\bf v}_n^{(i)}$ a right singular vector and ${\bf u}_n^{(i)}$ a left singular vector. It is straightforward to show that the corresponding generator vectors are mutually orthogonal. Note that ${\bf u}_n^{(2)}$ may be undefined for $n>C$ when $C<N$, but then $s_n^{(2)}=0$ and the block vanishes. The $N^2$ orthogonal generators span the null eigenspace of $\sf H$, establishing exactly $N^2$ zero eigenvalues. The generalization to arbitrary input, hidden, and output dimensions, including the bottleneck regimes where naive parameter counting fails, is given in the SM~\cite{supp}.

\setcounter{equation}{0}
\renewcommand{\theequation}{C\arabic{equation}}

\emph{Appendix C: $\varepsilon^2$ scaling and confinement of the bulk---}We make precise the two statements used in the main text. Write the Leaky-ReLU activation as
\begin{align}
    g^{(\varepsilon)}(z) = \big(1-\tfrac{\varepsilon}{2}\big)\,z + \tfrac{\varepsilon}{2}\,|z| \ ,
\end{align}
so that the two-layer network splits into a linear and an absolute-value part,
\begin{align}
    \bm f_{\rm LReLU}^{(\varepsilon)}(\boldsymbol{\theta},{\bf x}) = \big(1-\tfrac{\varepsilon}{2}\big) \bm  f_{\rm lin}(\boldsymbol{\theta},{\bf x}) + \tfrac{\varepsilon}{2}\, \bm f_{|\cdot|}(\boldsymbol{\theta},{\bf x}) \ .
    \label{eq:f_split}
\end{align}
At the interpolating minimum ${\sf H}^{(\varepsilon)}\equiv{\sf F}^{(\varepsilon)}=\langle (\partial_{\bm\theta^\top\!} \bm f^{(\varepsilon)})^\top \partial_{\bm\theta^\top\!} \bm f^{(\varepsilon)}\rangle_{\bf x}$. With ${\sf J}_0=\partial_{\bm\theta} \bm f_{\rm lin}^\top$ and ${\sf J}_1=\partial_{\bm\theta} \bm f_{|\cdot|}^\top$,
\begin{align}
    {\sf F}^{(\varepsilon)} &= \big(1-\tfrac{\varepsilon}{2}\big)^2\langle {\sf J}^{\vphantom{\top}}_0 {\sf J}_0^\top\rangle
    + \big(1-\tfrac{\varepsilon}{2}\big)\tfrac{\varepsilon}{2}\langle {\sf J}^{\vphantom{\top}}_0 {\sf J}^\top_1 + {\sf J}^{\vphantom{\top}}_1 {\sf J}^\top_0\rangle \nonumber\\
    &\quad + \big(\tfrac{\varepsilon}{2}\big)^2\langle {\sf J}_1^{\vphantom{\top}} {\sf J}_1^\top\rangle \ .
\end{align}
The cross term vanishes identically. Under ${\bf x}\to-{\bf x}$ the centered Gaussian measure is invariant, ${\sf J}_0({\bf x})$ is odd and ${\sf J}_1({\bf x})$ is even, so $\langle {\sf J}_0^{\vphantom{\top}} {\sf J}_1^\top\rangle=0$. The Fisher is therefore given by
\begin{align}
    {\sf F}^{(\varepsilon)} = \big(1-\tfrac{\varepsilon}{2}\big)^2\,{\sf F}_{\rm lin} + \big(\tfrac{\varepsilon}{2}\big)^2\,{\sf F}_{|\cdot|} \ ,
    \label{eq:fisher_pencil}
\end{align}
with ${\sf F}_{\rm lin}=\langle {\sf J}_0^{\vphantom{\top}} {\sf J}_0^\top\rangle$ and ${\sf F}_{|\cdot|}=\langle {\sf J}_1^{\vphantom{\top}} {\sf J}^\top_1\rangle$.

Let ${\sf P}$ project onto the symmetry subspace---the null space of ${\sf F}_{\rm lin}$, spanned by the $N^2$ generators---and ${\sf Q}={\sf 1}-{\sf P}$. Since ${\sf F}_{\rm lin}{\sf P}=0$, the symmetry block and the off-diagonal block are both $\mathcal{O}(\varepsilon^2)$,
\begin{align}
    \big({\sf F}^{(\varepsilon)}\big){\vphantom{\sf F}}^{\sf PP} = \big(\tfrac{\varepsilon}{2}\big)^2 {\sf F}^{\sf PP}_{|\cdot|}, \qquad
    \big({\sf F}^{(\varepsilon)}\big){\vphantom{\sf F}}^{\sf QP} = \big(\tfrac{\varepsilon}{2}\big)^2 {\sf F}^{\sf QP}_{|\cdot|} \ ,
\end{align}
while the curvature block $\big({\sf F}^{(\varepsilon)}\big){\vphantom{\sf F}}^{\sf QQ}=(1-\tfrac{\varepsilon}{2})^2{\sf F}_{\rm lin}^{\sf QQ}+\mathcal{O}(\varepsilon^2)$ stays $\mathcal{O}(1)$. Here, $(\cdot)^{\sf PP}\!\in\mathbb{R}^{{\rm rank}({\sf P})\times{\rm rank}({\sf P})}$, ${\rm rank}({\sf P})=N^2$,
denotes the block of a matrix restricted to the symmetry subspace, with analogous
restrictions $(\cdot)^{\sf QP}$, $(\cdot)^{\sf PQ}$ and $(\cdot)^{\sf QQ}$. Two conclusions follow.

\begin{figure}[t]
\centering
\includegraphics{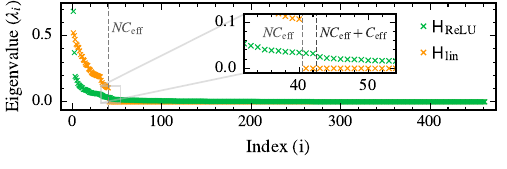}
\includegraphics{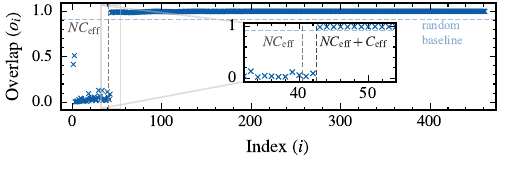}
\caption{Two-layer student--teacher model ($N=20$, $C=3$) with softmax output, single draw of Gaussian teacher weights ${\sf W}^{(k)}_{ij} \sim \mathcal{N}(0,1/N)$, Hessian at the global minimum. Softmax reduces the effective output width to $C_{\rm eff}=2$, giving $NC_{\rm eff}=40$ instead of $NC=60$ finite eigenvalues in the linear comparison network (top). The overlap $o_i$ (bottom) is near unity for the correspondingly increased number of small-eigenvalue ReLU modes.}
\label{fig:shallow_relu_model_softmax}
\end{figure}

\emph{(i) Scaling.} An eigenvalue branch leaving a degenerate level is analytic in the perturbation strength---here $(\varepsilon/2)^2$---so to leading order the bulk eigenvalues are $\lambda_i(\varepsilon)=\tfrac{\varepsilon^2}{4}\mu_i+\mathcal{O}(\varepsilon^4)$~\cite{Kato.1995}, with $\mu_i$ the eigenvalues of ${\sf F}^{\sf PP}_{|\cdot|}$. The naive expectation of a linear rise (from ${\sf F}^{(\varepsilon)}={\sf F}_{\rm lin}+\mathcal{O}(\varepsilon)$) is removed by the projection, which eliminates the linear term inside the symmetry block. Note that the $\mathcal{O}(\varepsilon^2)$ scaling of the eigenvalues persists even when the cross term $\langle {\sf J}_0^{\vphantom{\top}} {\sf J}_1^\top + {\sf J}_1^{\vphantom{\top}} {\sf J}_0^\top\rangle$ is nonzero---for non-Gaussian data, more than two weight layers, or different activation functions---since ${\sf P}{\sf J}_0^{\vphantom{\top}}=0$ still holds. Due to the positive homogeneity of Leaky-ReLU, the $N$ diagonal-rescaling generators remain exact symmetries and stay pinned at zero for all $\varepsilon$.

\emph{(ii) Confinement.} The second-order coefficient of the eigenvalue branch leaving the degenerate level is given by an inter-block term~\cite{Kato.1995}
\begin{align}
    \lambda_i(\varepsilon) = \tfrac{\varepsilon^2}{4}\mu_i
    - \tfrac{\varepsilon^4}{16}\,\big\lVert ({\sf F}_{\rm lin}^{\sf QQ})^{-1/2}{\sf F}^{\sf QP}_{|\cdot|}\,{\bf p}_{{\sf P},i}\big\rVert^2 + \mathcal{O}(\varepsilon^6) \ ,
    \label{eq:branch}
\end{align}
where ${\bf p}_{{\sf P},i}\in \mathbb{R}^{{\rm rank}({\sf P})}$ is the corresponding eigenvector of ${\sf F}^{\sf PP}_{|\cdot|}$. The $\varepsilon^4$ coefficient is a non-negative quadratic form in the leakage ${\sf F}^{\sf QP}_{|\cdot|}\,{\bf p}_{{\sf P},i}$ (the curvature block ${\sf F}_{\rm lin}^{\sf QQ}$ is positive definite), and vanishes \emph{if and only if} the leakage is zero, i.e.\ iff the mode is confined to the symmetry subspace. Hence a bulk eigenvalue that follows $\varepsilon^2$ with no $\varepsilon^4$ correction signifies a confined eigenvector. The empirical $\varepsilon^2$ trend across the bulk in \cref{fig:epsilon_scaling}, up to $\varepsilon=1$, is the spectral counterpart of the direct overlap measurement in \cref{fig:shallow_relu_model}. The $C$ modes bending away are exactly those with significant leakage---the mean-gradient outliers.

\emph{Appendix D: Softmax and the effective output dimension---}For a linear classifier the number of finite Hessian eigenvalues is given by the input dimension $N$ multiplied by the output dimension $C$. A softmax output activation normalizes the outputs and is invariant under uniform shifts of the $C$ preactivations~\cite{Kunin.2021}, an exact symmetry that reduces the effective output dimension to $C_{\rm eff}=C-1$ and the finite-eigenvalue count to $NC_{\rm eff}$. This exact rank reduction is distinct from the overall curvature suppression of softmax cross-entropy at high confidence late in training~\cite{Cohen.2021}. \cref{fig:shallow_relu_model_softmax} shows the effect in a modified version of the two-layer student--teacher model of \cref{eq:student_teacher_loss_lrelu}: with softmax, the linear network has $NC_{\rm eff}$ finite eigenvalues, the ReLU network has $\approx NC_{\rm eff}$ large modes with the rest at near-unity overlap, and the number of mean-gradient outliers is $C_{\rm eff}$ rather than $C$. 
For the two-layer mean-squared error model of \cref{eq:student_teacher_loss_lrelu}, the Hessian is obtained analytically by evaluating the Gaussian integrals over the input distribution in closed form. Here, with softmax and cross-entropy, we approximate the Hessian using a finite sample of $32{,}768$ Gaussian inputs. After sampling, the empirical mean is subtracted and the covariance is normalized to the identity, so that the finite dataset reproduces ${\bf x}\sim\mathcal{N}(0,{\sf 1}_{N})$ in its first two moments.

\ifarXiv
    

\section*{Supplementary material}

# Supplemental Material for: "Explaining Near-Zero Hessian Eigenvalues Through Approximate Symmetries in Neural Networks"

Marcel Kühn[1,2] and Bernd Rosenow[1]

[1]Institut für Theoretische Physik, Universität Leipzig, 04103 Leipzig, Germany
[2]ScaDS.AI Dresden/Leipzig, Universität Leipzig, 04105 Leipzig, Germany

## I. SYMMETRIES IN LINEAR FULLY CONNECTED LAYERS WITH ARBITRARY LAYER SIZE

In the main text we demonstrated, for a two-layer linear network with equal input and hidden dimension, that architectural symmetries generate exactly as many independent directions as there are zero eigenvalues of the Hessian, thereby fully explaining the null subspace of the spectrum. Here we extend this analysis to two consecutive linear fully connected layers with arbitrary input, hidden, and output dimensions. The discussion applies equally to any such pair of linear layers embedded in larger architectures, including deeper multilayer networks or linear blocks in more complex models.

We consider the linear map

$$\boldsymbol{f}_{\text{lin}}^{CMN}(\boldsymbol{\theta}, \mathbf{x}) = \mathsf{W}^{(2)}\mathsf{W}^{(1)}\mathbf{x} \ , \quad \boldsymbol{\theta} = \begin{pmatrix} \text{vec}\big(\mathsf{W}^{(1)}\big) \\ \text{vec}\big(\mathsf{W}^{(2)}\big) \end{pmatrix} \in \mathbb{R}^d \tag{S1}$$

with weights $\mathsf{W}^{(1)} \in \mathbb{R}^{M\times N}$ and $\mathsf{W}^{(2)} \in \mathbb{R}^{C\times M}$. The parameter vector has dimension $d = M(N + C)$.

Bias terms for the first layer may be incorporated straightforwardly by augmenting the input with a constant component. For the two-layer affine linear network $\boldsymbol{f}_{\text{aff}}(\mathbf{x}) = \mathsf{W}^{(2)}\big(\mathsf{W}^{(1)}\mathbf{x} + \mathbf{b}^{(1)}\big) + \mathbf{b}^{(2)}$, define $\tilde{\mathbf{x}} := \big(\mathbf{x}; 1\big)$, $\tilde{\mathsf{W}}^{(1)} := \big(\mathsf{W}^{(1)} \ \ \mathbf{b}^{(1)}\big)$, so that $\boldsymbol{f}_{\text{aff}}(\mathbf{x}) = \mathsf{W}^{(2)}\tilde{\mathsf{W}}^{(1)}\tilde{\mathbf{x}} + \mathbf{b}^{(2)}$. The hidden-layer transformation therefore acts on $\tilde{\mathsf{W}}^{(1)}$ and $\mathsf{W}^{(2)}$ while the output bias $\mathbf{b}^{(2)}$ is unchanged. Equivalently, first-layer biases transform together with the incoming weights. A complete count of the affine redundancies must also include the bias translation

$$\mathbf{b}^{(1)} \mapsto \mathbf{b}^{(1)} + \mathbf{u}, \qquad \mathbf{b}^{(2)} \mapsto \mathbf{b}^{(2)} - \mathsf{W}^{(2)}\mathbf{u} \qquad (\mathbf{u} \in \mathbb{R}^M), \tag{S2}$$

which shifts the offset entering the hidden layer while leaving the affine map $\boldsymbol{f}_{\text{aff}}$ invariant. These $M$ additional zero modes, together with the interlayer weight symmetries and the trainable output bias itself, leave generically $NC + C$ finite eigenvalues, in agreement with the main text. In the remainder of this section we suppress biases to keep the notation uncluttered.

Since the input-output relation can always be represented by a single effective matrix in $\mathbb{R}^{C\times N}$, the model is generically overparametrized. Away from bottlenecks, one therefore expects $d - NC = M(N + C) - NC$ independent symmetry directions. Specifically, we seek generators $\boldsymbol{\phi}'(\boldsymbol{\theta})$ such that

$$\boldsymbol{f}_{\text{lin}}^{CMN}(\boldsymbol{\theta} + \epsilon\boldsymbol{\phi}'(\boldsymbol{\theta}), \mathbf{x}) = \boldsymbol{f}_{\text{lin}}^{CMN}(\boldsymbol{\theta}, \mathbf{x}) + \mathcal{O}(\epsilon^2) \ , \tag{S3}$$

which provide zero-eigenvalue directions of the Hessian. As in the main text, these generators are induced by the algebra of linear transformations acting in the $M$-dimensional hidden layer. For any $\mathsf{A} \in \mathbb{R}^{M\times M}$, the corresponding generator is

$$\boldsymbol{\phi}'_{\mathsf{A}} = \begin{pmatrix} \text{vec}(\mathsf{A}\mathsf{W}^{(1)}) \\ \text{vec}(-\mathsf{W}^{(2)}\mathsf{A}) \end{pmatrix} \ . \tag{S4}$$

To obtain a mutually orthogonal set of generators, we again employ the singular value decompositions, having $\mathsf{W}^{(1)} = \mathsf{U}^{(1)}\mathsf{S}^{(1)}\big(\mathsf{V}^{(1)}\big)^\top$ with the orthogonal matrices $\mathsf{U}^{(1)} = \big(\mathbf{u}_1^{(1)}, \ldots, \mathbf{u}_M^{(1)}\big) \in \mathbb{R}^{M\times M}$ and $\mathsf{V}^{(1)} = \big(\mathbf{v}_1^{(1)}, \ldots, \mathbf{v}_N^{(1)}\big) \in \mathbb{R}^{N\times N}$ consisting of the left and right singular vectors. The rectangular matrix $\mathsf{S}^{(1)} = \text{diag}\big(s_1^{(1)}, \ldots, s_{\min(M,N)}^{(1)}\big) \in \mathbb{R}^{M\times N}$ contains the singular values on the main diagonal. An analogous structure holds for $\mathsf{W}^{(2)}$. We define a basis of $\mathbb{R}^{M\times M}$ by $\mathsf{A}^{(k,n)} := \mathbf{v}_k^{(2)}\big(\mathbf{u}_n^{(1)}\big)^\top$ which yields

$$\mathsf{A}^{(k,n)}\mathsf{W}^{(1)} = s_n^{(1)}\,\mathbf{v}_k^{(2)}\big(\mathbf{v}_n^{(1)}\big)^\top, \tag{S5a}$$

$$\mathsf{W}^{(2)}\mathsf{A}^{(k,n)} = s_k^{(2)}\,\mathbf{u}_k^{(2)}\big(\mathbf{u}_n^{(1)}\big)^\top \ . \tag{S5b}$$

When written as vectors, the corresponding generators take the form

$$\phi'^{(k,n)}_{CMN} \propto \begin{pmatrix} s_n^{(1)}\, v_{k,1}^{(2)} \mathbf{v}_n^{(1)} \\ \vdots \\ s_n^{(1)}\, v_{k,M}^{(2)} \mathbf{v}_n^{(1)} \\ -s_k^{(2)}\, u_{k,1}^{(2)} \mathbf{u}_n^{(1)} \\ \vdots \\ -s_k^{(2)}\, u_{k,C}^{(2)} \mathbf{u}_n^{(1)} \end{pmatrix} \in \mathbb{R}^{CM+MN} \ . \tag{S6}$$

Owing to the orthogonality of the singular vectors, these generators satisfy $\phi'^{(k,n)} \cdot \phi'^{(q,r)} \propto \delta_{kq}\delta_{nr}$. Therefore, in the case $N = M$, Eq. (S6) defines $N^2$ different orthogonal generators forming a basis for the null eigenspace of the Hessian as described in the main text. In general one must treat the different configurations of $C$, $M$, and $N$ more carefully. Note that when $M > N$ or $M > C$, the vectors $\mathbf{v}_n^{(1)}$ for $n > N$ or $\mathbf{u}_k^{(2)}$ for $k > C$ are strictly speaking undefined. However, in these cases the associated singular values vanish, so the corresponding vector components are set identically to zero.

We distinguish the relevant dimensional regimes:

*1.* $M \le \min(N, C)$. For all $(k, n) \in \{1, \dots, M\}^2$, at least one of the two blocks in Eq. (S6) is nonzero. Hence all $M^2$ generators are finite and linearly independent, yielding exactly $M^2$ symmetries. This holds in particular in bottleneck configurations, where the naive parameter-counting argument $(d - NC)$ would underestimate the number of symmetry directions $(M^2)$.

*2.* $M \ge \max(N, C)$. In this case one would naively obtain $M^2$ generators. However, only $M^2-(M-N)(M-C) = MN+MC-NC$ of them are independent. Indeed, for $(k, n)$ such that both $s_n^{(1)} = 0$ and $s_k^{(2)} = 0$, the vector Eq. (S6) vanishes. Removing these $(M-N)(M-C)$ zero vectors leaves precisely the expected number of symmetry generators.

*3.* $C < M < N$ *(and analogously* $N < M < C$*).* Here the construction above yields $M^2$ generators, while parameter counting predicts an additional $(N - M)(M - C)$ symmetry directions. These arise from additional generators that emerge from our definition via the singular value decomposition. For indices $n > M$ and $k > C$, additional generators are given by

$$\phi'^{(k>C,n>M)}_{C<M<N} \propto \begin{pmatrix} v_{k,1}^{(2)} \mathbf{v}_n^{(1)} \\ \vdots \\ v_{k,M}^{(2)} \mathbf{v}_n^{(1)} \\ 0 \\ \vdots \\ 0 \end{pmatrix} \ . \tag{S7}$$

These vectors are well-defined since $\mathbf{v}_n^{(1)}$ exists for $n \le N$, are orthogonal to all previously constructed generators, and satisfy Eq. (S3) because $\mathsf{W}^{(2)}\mathbf{v}_k^{(2)} = 0$ for $k > C$. Together with the $M^2$ generators above, they provide exactly $MN + MC - NC$ independent symmetry directions.

For more than two linear layers, dependencies between the different inter-layer generators can arise, depending in particular on the smallest layer width within the network, the bottleneck. A consistent interface-wise count is obtained by anchoring the count at a bottleneck. Consider a network with $L$ weight matrices, where matrix $l$ has dimension $M_{l-1} \times M_l$, with $M_0, ..., M_L \in \mathbb{N}$ and $N \equiv M_0$, $C \equiv M_L$. Let $q = \min(M_0, ..., M_L)$. It is known that such a linear map has $d_{\text{finite}} = q(N + C - q)$ finite Fisher modes [S1]. This number can also be derived by appropriate counting at interfaces: define the first bottleneck $b = \min\{j : M_j = q\}$. For an interface $i$ with hidden width $M_i$, one applies the two-layer rectangular count introduced above, which can be compactly defined as $Z(N, M, C) \coloneqq M(N + C) - \rho(N + C - \rho)$ with $\rho \coloneqq \min(N, M, C)$. However, one uses an effective dimension on the side pointing toward the bottleneck:

$$(N_i^{\text{eff}}, C_i^{\text{eff}}) = \begin{cases} (M_{i-1}, q)\,, & i \le b\ , \\ (M_{i-1}, M_{i+1})\,, & i = b\ , \\ (q, M_{i+1})\,, & i > b\ . \end{cases} \tag{S8}$$

Thus the total number of independent symmetry directions is $d_{\text{zero}} = \sum_{i=1}^{L-1} Z(N_i^{\text{eff}}, M_i, C_i^{\text{eff}})$. This gives $d_{\text{zero}} = \sum_{\ell=1}^{L} M_{\ell-1}M_\ell - q(M_0 + M_L - q)$, and hence the number of finite Fisher modes is $d_{\text{finite}} = q(N + C - q)$ or simply $NC$ in the case without bottlenecks.

In summary, for arbitrary input, hidden, and output dimensions, the formalism yields precisely the number of orthogonal generators required to account for all symmetry-induced zero eigenvalues of the Hessian. Throughout we assumed full rank of the weight matrices, as is generic at random initialization. If additional singular values vanish, further zero modes may appear, associated with generators analogous to those constructed in case 3.

## II. PERTURBATIVE EXPANSION OF THE FISHER AROUND LINEARITY AND THE ORIGIN OF RELU OUTLIERS

This section provides the technical basis for the perturbative treatment of nonlinear activations used in the main text, and for the origin of the large-eigenvalue outliers of the ReLU spectrum. We establish two statements. First, for *any* fully connected network built from the general activation deformation introduced below, the Fisher information matrix departs from its linear-network value only at order $\varepsilon$. Second, the bulk of the spectrum—exactly degenerate at zero in the linear network—is nonetheless lifted only at order $\varepsilon^2$, notwithstanding the $\mathcal{O}(\varepsilon)$ perturbation. We then leave the general formulation and specialize to a two-layer ReLU network, where we show analytically how the strict positivity of ReLU generates a dominant eigenvector aligned with the mean gradient in the large-$N$ limit.

Throughout we work at an interpolating minimum of the student–teacher loss, where the residuals vanish and the Fisher matrix $\mathsf{F}$ coincides with the Hessian $\mathsf{H}$. The identity $\mathsf{F} = \mathsf{H}$ holds equally for every variant appearing below, $\mathsf{F}_{\text{lin}} = \mathsf{H}_{\text{lin}}$, $\mathsf{F}^{(\varepsilon)} = \mathsf{H}^{(\varepsilon)}$, $\mathsf{F}_{\text{ReLU}} = \mathsf{H}_{\text{ReLU}}$.

### A. Perturbative Expansion of the Fisher Around the Linear Network

Any activation function $g_{\text{gen}}(x)$ can be written as a perturbation of the linear activation through the one-parameter family

$$g^{(\varepsilon)}_{\text{gen}}(x) := x + \varepsilon\big(g_{\text{gen}}(x) - x\big) \ , \tag{S9}$$

with $g^{(1)}_{\text{gen}}(x) = g_{\text{gen}}(x)$ and $g^{(\varepsilon)}_{\text{gen}}(x) = x + \mathcal{O}(\varepsilon)$. The construction is particularly natural for activations that are partly linear, such as ReLU—for which Eq. (S9) reduces to Leaky-ReLU with leakage $(1-\varepsilon)$—or for smooth activations, like the error function, that behave linearly near the origin, in which case one absorbs the appropriate Taylor coefficient into the linear term.

Consider a fully connected network of depth $L$ with output $\boldsymbol{f}^{(\varepsilon)}_{\text{FC}}(\mathbf{x})$ and activation $g^{(\varepsilon)}_{\text{gen}}$ applied entrywise in each hidden layer,

$$\boldsymbol{f}^{(\varepsilon)}_{\text{FC}}(\mathbf{x}) = \mathsf{W}^{(L)} g^{(\varepsilon)}_{\text{gen}}\Big(\mathsf{W}^{(L-1)} \cdots g^{(\varepsilon)}_{\text{gen}}\big(\mathsf{W}^{(1)}\mathbf{x}\big)\Big) \ . \tag{S10}$$

Because every activation obeys $g^{(\varepsilon)}_{\text{gen}} = \text{id} + \mathcal{O}(\varepsilon)$, replacing the activations by the identity turns Eq. (S10) into the multilayer linear network $\boldsymbol{f}_{\text{lin}}(\mathbf{x}) = \mathsf{W}^{(L)} \cdots \mathsf{W}^{(1)}\mathbf{x}$ with identical architecture and parameters. Expanding the composition, each nonlinearity contributes a correction of order $\varepsilon$, so the network output splits as

$$\boldsymbol{f}^{(\varepsilon)}_{\text{FC}}(\mathbf{x}) = \boldsymbol{f}_{\text{lin}}(\mathbf{x}) + \mathcal{O}(\varepsilon) \ . \tag{S11}$$

Differentiating with respect to the parameters, the Jacobian inherits the same decomposition,

$$\mathsf{J}^{(\varepsilon)} := \partial_{\boldsymbol{\theta}}\big(\boldsymbol{f}^{(\varepsilon)}_{\text{FC}}\big)^{\top} = \mathsf{J}_0 + \mathcal{O}(\varepsilon) \ , \qquad \mathsf{J}_0 = \partial_{\boldsymbol{\theta}} \boldsymbol{f}^{\top}_{\text{lin}} \ , \tag{S12}$$

with $\mathsf{J}_0$ the Jacobian of the linear network. The nondifferentiability of ReLU-type activations at the origin poses no obstruction here. For a finite dataset and generic parameters all preactivations are nonzero with probability one, so the network is a smooth, piecewise-polynomial function of the parameters in a neighborhood of $\boldsymbol{\theta}$. For the Gaussian population averages considered below, the set of inputs producing a vanishing preactivation for a generically nonzero incoming weight vector has measure zero, so Eq. (S12) holds almost everywhere and the population Jacobian is well defined.

At the interpolating minimum the Fisher is the input average of the Gauss–Newton form,

$$\mathsf{F}^{(\varepsilon)} = \big\langle \mathsf{J}^{(\varepsilon)} \mathsf{L}^{(\varepsilon)} \big(\mathsf{J}^{(\varepsilon)}\big)^{\top} \big\rangle_{\mathbf{x}} \ , \qquad \mathsf{L}^{(\varepsilon)} = \partial^2_{\boldsymbol{f}}\, \ell\big|_{\boldsymbol{f}=\boldsymbol{f}^{(\varepsilon)}_{\text{FC}}} \ , \tag{S13}$$

where $\mathsf{L}^{(\varepsilon)}$ is the metric induced on output space by the per-example loss $\ell$. For the mean-squared error $\ell = \frac{1}{2}\|\boldsymbol{f} - \boldsymbol{\tau}\|^2$ this metric is the constant identity, $\mathsf{L}^{(\varepsilon)} = 1_C$, so the output-space factor drops out and

$$\mathsf{F}^{(\varepsilon)} = \left\langle \mathsf{J}^{(\varepsilon)} \big(\mathsf{J}^{(\varepsilon)}\big)^\top \right\rangle_{\mathbf{x}} = \left\langle \mathsf{J}_0 \mathsf{J}_0^\top \right\rangle_{\mathbf{x}} + \mathcal{O}(\varepsilon) = \mathsf{F}_{\text{lin}} + \mathcal{O}(\varepsilon) \ , \tag{S14}$$

with $\mathsf{F}_{\text{lin}} = \langle \mathsf{J}_0 \mathsf{J}_0^\top \rangle_{\mathbf{x}}$ the Fisher of the linear network. This establishes the expansion in $\varepsilon$ claimed in the main text, now for an arbitrary fully connected network rather than the two-layer case alone. The argument is not specific to the MSE: for a softmax read-out with cross-entropy loss the metric is $\mathsf{L}^{(\varepsilon)} = \text{diag}(\mathbf{p}) - \mathbf{p}\mathbf{p}^\top$ with $\mathbf{p} = \text{softmax}(\boldsymbol{f}_{\text{FC}}^{(\varepsilon)})$; since $\boldsymbol{f}_{\text{FC}}^{(\varepsilon)} = \boldsymbol{f}_{\text{lin}} + \mathcal{O}(\varepsilon)$ by Eq. (S11), the metric itself is $\mathsf{L}^{(\varepsilon)} = \mathsf{L}_{\text{lin}} + \mathcal{O}(\varepsilon)$ and bounded, so Eq. (S13) again yields $\mathsf{F}^{(\varepsilon)} = \mathsf{F}_{\text{lin}} + \mathcal{O}(\varepsilon)$.

### B. Scaling of the Lifted Bulk

The expansion Eq. (S14) would naively suggest that eigenvalues degenerate at zero in the linear network rise linearly in $\varepsilon$. They do not: the bulk is lifted only at order $\varepsilon^2$. We now make this precise. Let $\mathsf{P}$ project onto the symmetry subspace—the null space of $\mathsf{F}_{\text{lin}}$, spanned by the function-preserving generators—with $\mathsf{Q} = 1 - \mathsf{P}$.

The key structural fact is that the linear Jacobian is annihilated by $\mathsf{P}$,

$$\mathsf{P}\,\mathsf{J}_0 = 0 \ . \tag{S15}$$

Geometrically, the generators of a function-preserving symmetry leave the network output unchanged, so the gradient of the linear network is orthogonal to them. Consequently the projected Jacobian of the full network starts at first order,

$$\mathsf{P}\,\mathsf{J}^{(\varepsilon)} = \mathsf{P}\,\mathsf{J}_0 + \mathcal{O}(\varepsilon) = \mathcal{O}(\varepsilon) \ . \tag{S16}$$

The Fisher is *quadratic* in the Jacobian, so every block that carries a factor of $\mathsf{P}$ inherits the corresponding power of $\varepsilon$, resulting in

$$\big(\mathsf{F}^{(\varepsilon)}\big)^{\mathsf{PP}} = \mathcal{O}(\varepsilon^2) \ . \tag{S17}$$

Here, $(\cdot)^{\mathsf{PP}} \in \mathbb{R}^{\text{rank}(\mathsf{P}) \times \text{rank}(\mathsf{P})}$ denotes the block of a matrix restricted to the symmetry subspace.

An eigenvalue branch leaving a degenerate level is analytic in the perturbation strength [S2], so each bulk eigenvalue has a well-defined leading power of $\varepsilon$. Moreover, for any vector $\mathbf{v}$ in the symmetry subspace, i.e. $\mathbf{v} \in \text{ran}(\mathsf{P})$, one has $\mathbf{v}^\top \mathsf{F}^{(\varepsilon)} \mathbf{v} = \mathcal{O}(\varepsilon^2)\|\mathbf{v}\|^2$. By the Courant–Fischer–Weyl min-max principle, the rank$(\mathsf{P})$ smallest eigenvalues are therefore bounded above by $\big\|(\mathsf{F}^{(\varepsilon)})^{\mathsf{PP}}\big\| = \mathcal{O}(\varepsilon^2)$. Hence the leading power of $\varepsilon$ for each of the rank$(\mathsf{P})$ bulk eigenvalues is at least quadratic, $\lambda_i^{\text{bulk}}(\varepsilon) = \mathcal{O}(\varepsilon^2)$.

For the two-layer Gaussian model this general statement reduces to the closed form of the End Matter. The $\varepsilon^2$ onset of the bulk is robust beyond that two-layer model. Even for non-Gaussian data, more than two weight layers, or other activations, $\lambda_i^{\text{bulk}}(\varepsilon) = \mathcal{O}(\varepsilon^2)$ still holds, because $\mathsf{P}\mathsf{J}_0 = 0$ remains valid.

### C. Origin of the Large ReLU Outlier Eigenvalue

We now leave the general formulation and specialize to standard ReLU, in the concrete setting of a two-layer network, in order to expose the mechanism behind the mean-gradient outliers. The calculation below is intended as a tractable illustration of that mechanism, not as a general characterization of trained ReLU networks. We analytically demonstrate how the strict positivity of ReLU activations produces a dominant Fisher eigenvector aligned with the mean gradient in the large-$N$ limit. Consider a two-layer ReLU network with one output, $f_{\text{ReLU}}(\boldsymbol{\theta}, \mathbf{x}) := \sum_{n=1}^{N} \mathsf{W}_n^{(2)} \text{ReLU}(\sum_{m=1}^{N} \mathsf{W}_{nm}^{(1)} \mathsf{x}_m)$, with parameters $\boldsymbol{\theta} = \big(\text{vec}(\mathsf{W}^{(2)}); \text{vec}(\mathsf{W}^{(1)})\big) \in \mathbb{R}^{N+N^2}$, ordered to match the gradient below. We analyze the Fisher at the specific point $\boldsymbol{\theta}_0$ with $\mathsf{W}^{(1)} = 1_N$ and $\mathsf{W}^{(2)} = (1/\sqrt{N}, \ldots, 1/\sqrt{N})$, which by rotational symmetry is representative of any orthogonal $\mathsf{W}^{(1)}$.

For this choice, the gradient takes the explicit form $\partial_{\boldsymbol{\theta}} f_{\text{ReLU}}(\boldsymbol{\theta}_0, \mathbf{x}) = \big(\text{ReLU}(\mathbf{x});\ \frac{1}{\sqrt{N}}\mathbf{x}\Theta(\mathsf{x}_1);\ \ldots;\ \frac{1}{\sqrt{N}}\mathbf{x}\Theta(\mathsf{x}_N)\big)$, and the Fisher is given by

$$\mathsf{F}_{\text{ReLU}} = \langle \partial_{\boldsymbol{\theta}} f_{\text{ReLU}}(\boldsymbol{\theta}_0, \mathbf{x}) \partial_{\boldsymbol{\theta}^\top} f_{\text{ReLU}}(\boldsymbol{\theta}_0, \mathbf{x}) \rangle_{\mathbf{x}} \ . \tag{S18}$$

Its entries involve Gaussian averages of the form $\langle \mathrm{x}_i \Theta(\mathrm{x}_j) \mathrm{x}_n \Theta(\mathrm{x}_m)\rangle$, which are straightforward to evaluate.

Rather than diagonalizing the full $(N^2+N)\times(N^2+N)$ matrix, we project onto a three-dimensional subspace spanned by the nonorthonormal basis vectors

$$\begin{aligned} \mathbf{b}_1 &= (\mathbf{h};\ \mathbf{0};\quad \mathbf{0};\quad \ldots;\ \mathbf{0}) \\ \mathbf{b}_2 &= (\mathbf{0};\ \mathbf{h};\quad \mathbf{h};\quad \ldots;\ \mathbf{h}) \\ \mathbf{b}_3 &= (\mathbf{0};\ \mathbf{e}^{(1)};\ \mathbf{e}^{(2)};\ \ldots;\ \mathbf{e}^{(N)})\ , \end{aligned} \tag{S19}$$

where $\mathrm{h}_i = 1$ and $\mathrm{e}_j^{(i)} = \delta_{ij}$. Defining $\mathsf{B} = (\mathbf{b}_1, \mathbf{b}_2, \mathbf{b}_3)$, the projected Fisher $\tilde{\mathsf{F}}_{\mathrm{ReLU}} = (\mathsf{B}^\top \mathsf{B})^{-1}\mathsf{B}^\top \mathsf{F}_{\mathrm{ReLU}}\mathsf{B}$ is

$$\tilde{\mathsf{F}}_{\mathrm{ReLU}} = \frac{1}{4\pi}\begin{pmatrix} 2N + 2(\pi-1) & (2+\pi)\sqrt{N} + (\pi-2)\frac{1}{\sqrt{N}} & 2\sqrt{N} + 2(\pi-1)\frac{1}{\sqrt{N}} \\ \pi\frac{1}{\sqrt{N}} & \pi + (2+\pi)\frac{1}{N} & \pi\frac{1}{N} \\ 2\sqrt{N} + (\pi-2)\frac{1}{\sqrt{N}} & 2 - 4\frac{1}{N} & 2 + (\pi-2)\frac{1}{N} \end{pmatrix}\ . \tag{S20}$$

One verifies that $\mathsf{F}_{\mathrm{ReLU}}\mathbf{b}_i$ is always a linear combination of the $\mathbf{b}_i$, ensuring that eigenvectors of $\tilde{\mathsf{F}}_{\mathrm{ReLU}}$ lift to eigenvectors of the full Fisher. Diagonalization yields, for the largest eigenvalue, the eigenvector $\mathbf{p}_1 \propto [1 + \mathcal{O}(1/N)]\mathbf{b}_1 + \mathcal{O}(1/N)\mathbf{b}_2 + \frac{1}{\sqrt{N}}\mathbf{b}_3$. In contrast, the mean gradient evaluates to $\langle \partial_{\boldsymbol{\theta}} f_{\mathrm{ReLU}}(\boldsymbol{\theta}_0, \mathbf{x})\rangle_{\mathbf{x}} = \frac{1}{\sqrt{2\pi}}\left(\mathbf{b}_1 + \frac{1}{\sqrt{N}}\mathbf{b}_3\right)$, demonstrating that the leading ReLU eigenvector aligns with the mean gradient up to $1/N$ corrections.

For the corresponding linear network with the same weights, the gradient is $\partial_{\boldsymbol{\theta}} f_{\mathrm{lin}}(\boldsymbol{\theta}_0, \mathbf{x}) = \left(\mathbf{x};\ \frac{1}{\sqrt{N}}\mathbf{x};\ \ldots;\ \frac{1}{\sqrt{N}}\mathbf{x}\right) \in \mathbb{R}^{N^2+N}$, which yields a block-structured Fisher $\mathsf{F}_{\mathrm{lin}}$. Its eigenvectors associated with finite eigenvalues are therefore given by $\mathbf{p}_j^{(\mathrm{lin})} \propto \left(\mathbf{e}^{(j)};\ \frac{1}{\sqrt{N}}\mathbf{e}^{(j)};\ \ldots;\ \frac{1}{\sqrt{N}}\mathbf{e}^{(j)}\right)$, while the remaining directions span the symmetry-induced zero-eigenvalue subspace. Using this structure and accounting for normalization, one finds in the large-$N$ limit that the overlap of the leading ReLU eigenvector $\mathbf{p}_1$ with the zero-eigenvalue subspace of the linear model approaches $o_1 = 0.5$. Thus, despite being associated with the largest eigenvalue of $\mathsf{F}_{\mathrm{ReLU}}$, $\mathbf{p}_1$ retains a substantial component along the linear symmetry directions due to the strict positivity of the ReLU activation. We expect this partial alignment to persist for more general choices of $\mathsf{W}^{(1)}$ and $\mathsf{W}^{(2)}$, providing a direct explanation for the outlier behavior observed in the overlap spectrum in the main text.

## III. THREE-LAYER STUDENT–TEACHER NETWORK

To show that the mechanism is not tied to a single hidden layer, we repeat the Leaky-ReLU analysis for a three-layer student–teacher network, with student and teacher sharing the architecture

$$\boldsymbol{f}^{(\varepsilon)}(\mathbf{x}) = \mathsf{W}^{(3)}\, g^{(\varepsilon)}\big(\mathsf{W}^{(2)}\, g^{(\varepsilon)}\big(\mathsf{W}^{(1)}\mathbf{x}\big)\big)\ , \tag{S21}$$

with input dimension $N = 20$, two hidden layers of width 20, and $C = 3$ outputs, and with the Hessian evaluated at the global minimum.

The eigenvalue-versus-$\varepsilon$ trajectory (Fig. S1, bottom) shows the bulk detaching from zero as $\mathcal{O}(\varepsilon^2)$ in the linear limit. In contrast to the two-layer case, however, the clean $\varepsilon^2$ law no longer reaches the ReLU point. Eigenvalue branches bend away from the quadratic trend before $\varepsilon = 1$, signaling higher-order corrections and hence eigenvectors that leak partially out of the symmetry subspace.

At the ReLU point the overlap spectrum (Fig. S1, right) nonetheless retains the two-tier structure found in the two-layer and CIFAR-10 cases. The bulk sits at overlap near unity, while the high-curvature modes are clearly suppressed. The suppression is weaker than in the shallow network—the leading modes level off near $o_i \approx 0.2$ rather than close to zero—but the gap between the two groups remains unambiguous, so the same symmetry organization is evidently at work.

The two settings also differ in how the data average over $\mathbf{x} \sim \mathcal{N}(0, 1_N)$ is taken. For the two-layer model the Hessian is obtained analytically, carrying out the Gaussian integrals over the input distribution in closed form. The additional interlayer dependence renders these integrals intractable, so for the three-layer network we approximate the Hessian on a finite sample of 32,768 Gaussian inputs. After sampling, the empirical mean is subtracted and the covariance normalized to the identity, so that the finite dataset reproduces $\mathbf{x} \sim \mathcal{N}(0, 1_N)$ in its first two moments.

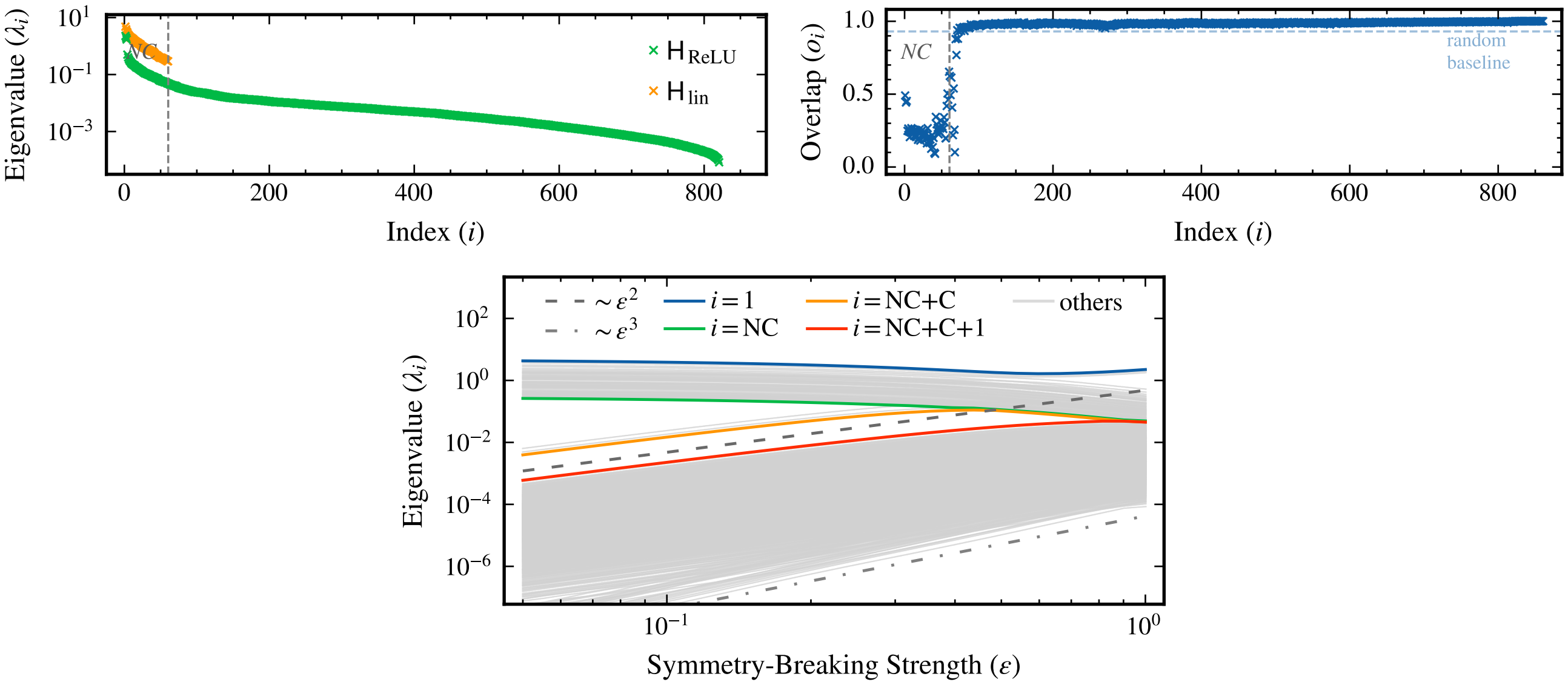


FIG. S1. Three-layer Leaky-ReLU student–teacher network (input dimension $N = 20$, two hidden layers of width 20, $C = 3$ outputs), Hessian at global minimum, with the data average approximated on 32,768 Gaussian samples. Left: ranked Hessian eigenvalues at the ReLU point. Right: overlap $o_i$ between the ReLU eigenvectors and the symmetry subspace of the linear comparison network; the two-tier structure persists—less completely than in the two-layer case, but still clearly separated. Bottom: ranked eigenvalues as a function of the symmetry-breaking strength $\varepsilon$. The bulk leaves zero as $\mathcal{O}(\varepsilon^2)$ near the linear limit but—unlike the two-layer model—most branches depart from the trend before the ReLU point $\varepsilon = 1$. The very smallest positive eigenvalues no longer scale as $\varepsilon^2$, for small $\varepsilon$ they instead appear to follow higher powers. Note that the rescaling modes again stay pinned at zero.

## IV. CIFAR-10 TRAINING AND DATA WHITENING

The three-layer ReLU network—3072-dimensional input, two hidden layers of 128 units, 10-way softmax, no biases, 410,880 parameters—is trained using stochastic gradient descent with momentum. We employ an initial learning rate of 0.005 together with an exponential decay schedule, reducing the learning rate by a factor 0.8 each epoch. Over 30 epochs this lowers the learning rate by approximately three orders of magnitude. The batch size is 32, the momentum factor is 0.9, weight decay $\lambda_{\mathrm{wd}} = 10^{-4}$ is applied, and all weights are initialized using the Glorot normal initializer. The learning rate and its schedule were selected by a grid search and yield near-optimal performance both for training on whitened data (defined below) and on data with its original variance, the latter producing a broader Hessian eigenspectrum. While training is stochastic, the Hessian is computed over the full training set.

The weight decay $\lambda_{\mathrm{wd}} = 10^{-4}$ is included in the optimization but *omitted* from the reported Hessian. An $L_2$ penalty $\lambda_{\mathrm{wd}}\|\boldsymbol{\theta}\|^2$ shifts every eigenvalue uniformly by $2\lambda_{\mathrm{wd}}$ while leaving all eigenvectors unchanged. It therefore plays no role in the overlap or coverage measurements, and including it would only add a constant to the eigenvalue axes.

Test accuracy for the whitened-data setting is $52.7\% \pm 0.4\%$ over three random seeds, while training on the original-variance input yields $53.3\% \pm 0.2\%$. These values agree with expectations for plain fully connected CIFAR-10 models [S3]. Whitening is a standard preprocessing technique associated with improved performance and faster convergence [S4, S5]. In our experiments the latter is evident from the fact that a test accuracy above 51% is reached after only three epochs when training on whitened data, whereas five epochs are required for inputs at their original variance.

Preprocessing begins by subtracting the mean image (computed over the training set) from each input. Let $\mathsf{X} = (\mathbf{x}_1, \ldots, \mathbf{x}_p)$ denote the data matrix, with $\mathbf{x}_i \in \mathbb{R}^N$, $N = 3072$. The sample covariance is

$$\Sigma_{\mathsf{X}} \coloneqq \frac{1}{p}\mathsf{X}\mathsf{X}^\top, \tag{S22}$$

with eigendecomposition $\Sigma_{\mathsf{X}} = \mathsf{O}\mathsf{D}\mathsf{O}^\top$, where $\mathsf{O}$ is orthogonal and $\mathsf{D} = \mathrm{diag}(\sigma_1^2, \ldots, \sigma_N^2)$ with $\sigma_i^2 \geq \sigma_{i+1}^2$. Full whitening would normalize all variances to unity, but amplifying very low-variance directions could strengthen noise in the data and is generally detrimental. Instead, we apply regularized whitening and normalize only the $N_W = 100$ largest variance directions, which together account for 90% of the total variance. To avoid amplifying small-variance

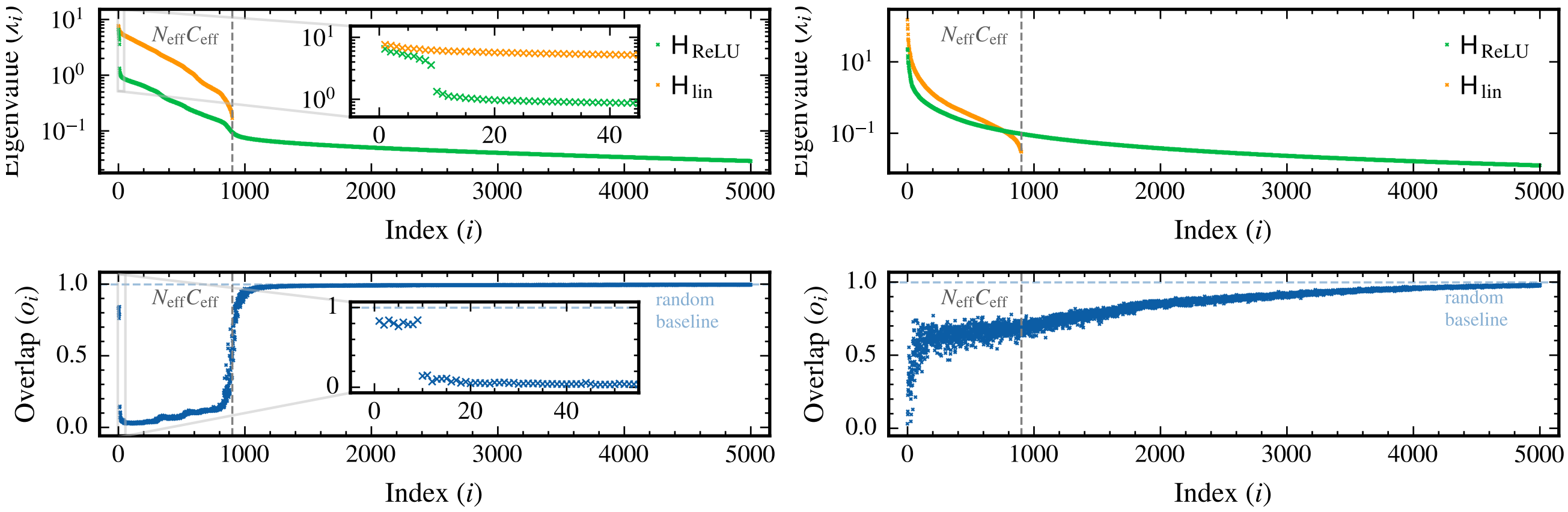


FIG. S2. Hessian eigenspectrum for the 5,000 largest eigenvalues and symmetry-overlap structure for CIFAR-10. Left: whitened inputs (normalized top-variance directions). Right: original data variance. The linear-network eigenvectors used to form the zero-mode comparison subspace are the same for both cases and obtained from a reduced model retaining only the 100 highest-variance input directions, implemented by setting all remaining variances to zero. The finite linear-network eigenvalues differ, as they are computed using the corresponding reduced data covariance. In the original-variance case the broader covariance spectrum increases the spread of Hessian eigenvalues and interacts with the symmetry mechanism, producing a smoother overlap profile. Nevertheless, eigenvectors associated with large Hessian eigenvalues continue to exhibit substantially reduced overlap with the symmetry directions.

components, we define the average scale $\overline{\sigma_W^2} = \frac{1}{N_W}\sum_{i=1}^{N_W}\sigma_i^2$ and normalize the full variance by this factor. Only then the subspace of the largest variance directions is decorrelated, leading to the regularized diagonal matrix

$$\mathsf{D}_W = \operatorname{diag}\left(\sigma_1^2, \dots, \sigma_{N_W}^2, \overline{\sigma_W^2}, \dots, \overline{\sigma_W^2}\right). \tag{S23}$$

and the whitening transform is then

$$\mathsf{V}_W = \mathsf{O}\,\mathsf{D}_W^{-1/2}\,\mathsf{O}^\top. \tag{S24}$$

The whitened data are $\mathsf{X}_W = \mathsf{V}_W\mathsf{X}$. This procedure leaves the top $N_W = 100$ directions with unit variance, still containing 90% of the cumulative variance, while the remaining directions retain significantly smaller variance, effectively yielding an input dimension $N_{\text{eff}} = 100$.

The connection between the data covariance $\Sigma_{\mathsf{X}}$ and the Hessian $\mathsf{H}$ is most straightforward in linear regression, where $\mathsf{H} = \Sigma_{\mathsf{X}}$ and a broad covariance spectrum directly produces a broad Hessian spectrum. In multilayer networks analogous effects persist and interact with symmetry-induced small Hessian eigenvalues. As described in the main text, for the linear comparison model we copy the trained ReLU parameters and remove the hidden nonlinearities but keep the softmax, giving an effective output dimension $C_{\text{eff}} := C - 1 = 9$ (End Matter). The comparison subspace, collectively termed the symmetry subspace $\mathsf{P}$, is the null space of the linear model's Fisher with the low-variance directions set to zero variance. It combines the suppressed-variance directions with the architectural symmetry directions. To be explicit, the $d - NC_{\text{eff}} = 383{,}232$ architectural symmetries decompose into subgroups: $2 \times 128^2 = 32{,}768$ symmetries connected to the general linear group, $(N - 128)(128 - C_{\text{eff}}) = 350{,}336$ symmetries related to directions of the first weight matrix that map onto the kernel of the downstream map, and 128 further symmetries in the last layer related to the softmax invariance. The remaining $(N - N_{\text{eff}})C_{\text{eff}} = 26{,}748$ zero modes are connected to the low-variance directions.

As shown in Fig. S2 (left), for whitened data the overlap between Hessian eigenvectors and the zero-mode subspace of the linear comparison model, covering all but $N_{\text{eff}}C_{\text{eff}} = 900$ dimensions of the full parameter space, is near zero for the 900 largest eigenvalues, with a noticeable distinction in magnitude between eigenvalues just below and above this threshold. For data at the original variance, Fig. S2 (right), the overlaps remain substantially smaller than the random baseline but vary more smoothly due to the richer variance spectrum, even though in both cases the 100 largest variance directions contribute 90% of the cumulative variance. Note that the comparison subspace $\mathsf{P}$ is the same as the one used for the whitened data; in the original variance setup the comparison Fisher matrix of the linear network is determined again with variance directions beyond the largest 100 variances set to zero, while the training loss Hessian of the nonlinear network is determined with respect to the full original variance data. The eigenvalue curve likewise shows no sharp feature at the 900th eigenvalue. This indicates that symmetry effects and the broad data-variance-driven structure of $\mathsf{H}$ interact. Eigenvectors with large Hessian eigenvalues may exhibit substantial overlap

with a symmetry direction if they also arise from high-variance input directions; absent the variance contribution, their eigenvalues could be significantly smaller.

Consider also the random baseline for the overlap $o_i$. Let $\mathsf{P}$ denote the orthogonal projector onto the zero-eigenvalue subspace of the linear comparison model, and write $o(\mathbf{p}) = \|\mathsf{P}\mathbf{p}\|^2$. For a unit vector $\mathbf{p}$ drawn uniformly from $S^{d-1}$, isotropy gives $o_{\text{random}} := \mathbb{E}_{\mathbf{p}\sim S^{d-1}}[o(\mathbf{p})] = \text{rank}(\mathsf{P})/d$. In the CIFAR-10 experiment $d = 410{,}880$, while the finite-eigenvalue subspace of the linear comparison model has dimension $N_{\text{eff}}C_{\text{eff}} = 900$, so $o_{\text{random}} \approx 0.9978$. The leading CIFAR-10 modes therefore constitute the evidence for symmetry organization precisely because their overlaps fall far below this baseline. For the fraction $O_k$ of the high-curvature subspace spanned by the leading $k$ Hessian eigenvectors, the original-variance experiment yields $O_{5000}^{\text{orig}} \approx 0.88$, compared to $O_{5000}^{\text{white}} \approx 0.95$ for whitened data. The first 5000 modes still account for 88% of the 900 linear curvature directions, so the remaining $\sim 4 \times 10^5$ modes still share only the leftover 12%.

Numerically, only the 5,000 largest Hessian eigenvalues of the nonlinear network and their eigenvectors are computed. A full diagonalization is not possible due to resource constraints. Storing the dense $410{,}880\times 410{,}880$ Hessian in double precision alone would require memory in the terabyte range. Instead, the leading eigenpairs were computed matrix-free from Hessian-vector products using `scipy.sparse.linalg.eigsh` / ARPACK's implicitly restarted Lanczos method, with default parameters; implementation details are provided in the code repository [S6]. The overlap with the predicted flat directions is not measured directly; rather, the overlap with the 900 finite-eigenvalue eigenvectors of the linear network is computed, and the overlap with the complementary (symmetry) subspace is inferred from it.

## V. DETAILS OF THE TWO-LAYER CONVOLUTIONAL NETWORK

In the main text we considered the linear one-dimensional convolutional network

$$f_{\text{conv}}(\boldsymbol{\theta}, \mathbf{x}) = \left(\mathbf{w}^{(2)}\right)^{\top}\left(\mathbf{w}^{(1)} * \mathbf{x}\right), \tag{S25}$$

with kernel $\mathbf{w}^{(1)} \in \mathbb{R}^K$, readout weights $\mathbf{w}^{(2)} \in \mathbb{R}^N$, and parameters $\boldsymbol{\theta} = (\mathbf{w}^{(1)}; \mathbf{w}^{(2)})$ stacked columnwise. The first layer performs a convolution,

$$(\mathbf{w}^{(1)} * \mathbf{x})_i = \sum_{j=0}^{K-1} \mathrm{w}_{j+1}^{(1)} \mathrm{x}_{i+j}\,. \tag{S26}$$

The example is deliberately minimal. Extending the construction to multilayer CNNs with padding, pooling, and channel mixing requires identifying the corresponding architecture-specific approximate symmetries.

In this section, we demonstrate that the example model possesses $K$ continuous symmetries when employing periodic boundary conditions in the convolution and that these symmetries are not linear in the weights (in contrast to fully connected layers), and we discuss how they persist approximately in the nonperiodic case.

To apply the same approach as for fully connected layers, we make use of the fact that a one-dimensional convolution is a linear transformation and can therefore be represented by a matrix multiplication. The same holds for a two-dimensional convolution, implying that the arguments below directly extend to a network whose first layer is a 2D convolution, followed by flattening of the hidden representation and a dot product with a readout weight vector.

For periodic boundary conditions, defining $\mathrm{x}_{N+i} := \mathrm{x}_i$ in Eq. (S26) with $\mathbf{x} \in \mathbb{R}^N$, the convolution can be written as $\mathbf{w}^{(1)} * \mathbf{x} = \mathsf{C}(\mathbf{w}^{(1)})\mathbf{x}$, where $\mathsf{C}(\mathbf{w}^{(1)})$ is a circulant matrix. For a vector $\mathbf{v} \in \mathbb{R}^N$ we define the corresponding circulant matrix $\mathsf{C}(\mathbf{v}) \in \mathbb{R}^{N\times N}$ as

$$\mathsf{C}(\mathbf{v}) = \begin{pmatrix} \mathrm{v}_1 & \mathrm{v}_2 & \dots & \mathrm{v}_{N-1} & \mathrm{v}_N \\ \mathrm{v}_N & \mathrm{v}_1 & \dots & \mathrm{v}_{N-2} & \mathrm{v}_{N-1} \\ \vdots & \vdots & \ddots & \vdots & \vdots \\ \mathrm{v}_3 & \mathrm{v}_4 & \dots & \mathrm{v}_1 & \mathrm{v}_2 \\ \mathrm{v}_2 & \mathrm{v}_3 & \dots & \mathrm{v}_N & \mathrm{v}_1 \end{pmatrix}. \tag{S27}$$

For the lower-dimensional kernel $\mathbf{w}^{(1)}$, all diagonals beyond the first $K$ are defined to be zero.

To determine symmetries, we employ the same approach as in the main text for fully connected layers, inserting a matrix exponential $e^{\alpha\mathsf{A}}$ and its inverse between layers to generate symmetry transformations and their generators. A notable difference is that the transformed matrix $e^{\alpha\mathsf{A}}\mathsf{C}(\mathbf{w}^{(1)})$ must itself be circulant for the transformation to be valid. Infinitesimally, we therefore seek generators satisfying

$$\mathsf{A}\mathsf{C}(\mathbf{w}^{(1)}) = \mathsf{C}(\boldsymbol{\delta}\mathbf{w}^{(1)})\,, \tag{S28}$$

corresponding to a valid change of the kernel.

For a randomly chosen $\mathbf{w}^{(1)}$, the circulant matrix $\mathsf{C}(\mathbf{w}^{(1)})$ is almost always invertible, which can be verified via diagonalization using the discrete Fourier transform matrix. The set of weights for which $\mathsf{C}$ is not invertible has measure zero. Invertibility implies that the generator $\mathsf{A}$ must itself be circulant. Ideally, one would like $\mathsf{A}$ to be valid independently of $\mathbf{w}^{(1)}$. However, again using the discrete Fourier transform, one finds that the only fixed circulant matrix $\mathsf{A}$ which maps $\mathsf{C}(\mathbf{w}^{(1)})$ to another valid circulant matrix $\mathsf{C}(\boldsymbol{\delta}\mathbf{w}^{(1)})$, where $\boldsymbol{\delta}\mathbf{w}^{(1)} \in \mathbb{R}^K$, for all $\mathbf{w}^{(1)} \in \mathbb{R}^K$ is proportional to the identity, $\mathsf{A} \propto \mathbb{1}$, corresponding to the trivial symmetry of upscaling the first layer and downscaling the second.

To obtain the full set of symmetries, we therefore need to define generator matrices that depend explicitly on the kernel weights. Since $\mathbf{w}^{(1)}$ has only $K$ independent entries, only $K$ generators are required, which can be chosen as

$$\mathsf{A}_{\mathrm{conv}}(i, \mathbf{w}^{(1)}) = \mathsf{C}\left(\mathbf{e}^{(i)}\right) \mathsf{C}^{-1}\left(\mathbf{w}^{(1)}\right), \tag{S29}$$

where $\mathbf{e}^{(i)}$ denotes the $K$-dimensional unit vector with entry $i$ equal to one. The resulting generators are no longer linear in the network parameters, and consequently neither are the associated symmetry transformations. This motivates the more general notation for symmetry transformations beyond those linear in $\boldsymbol{\theta}$. The $K$ symmetry generators are given by

$$\boldsymbol{\phi}'_{\mathsf{A}_{\mathrm{conv}}(i,\mathbf{w}^{(1)})} = \left(\mathbf{e}^{(i)}; \, -\mathsf{A}^{\top}_{\mathrm{conv}}(i, \mathbf{w}^{(1)})\, \mathbf{w}^{(2)}\right). \tag{S30}$$

In the unlikely case that $\mathsf{C}(\mathbf{w}^{(1)})$ is not invertible and Eq. (S29) is ill-defined, symmetry generators still exist. For example, if $\mathbf{v}^{\top}\mathsf{C}(\tilde{\mathbf{w}}^{(1)}) = 0$, one may choose $\mathsf{A} \propto \mathbf{w}^{(2)}\mathbf{v}^{\top}$, yielding $\boldsymbol{\phi}'_{\mathsf{A}} = (0; \mathbf{v})$.

Having established these symmetry generators in the periodic case, which define zero-eigenvalue Hessian eigenvectors, we now ask whether the symmetries persist approximately when the exact symmetry is broken. To this end, we consider the more common case of nonperiodic boundary conditions and compare the symmetry generators of the periodic network, given by Eq. (S30), with the Hessian eigenvectors of the same network employing a nonperiodic convolution. Specifically, we study the Hessian of the student–teacher loss

$$\mathcal{L}_{\mathrm{conv}}(\boldsymbol{\theta}) = \frac{1}{2}\left\langle \left(f^{\mathrm{n.p.}}_{\mathrm{conv}}(\boldsymbol{\theta}, \mathbf{x}) - \tau_{\mathrm{conv}}(\mathbf{x})\right)^2 \right\rangle_{\mathbf{x}}, \tag{S31}$$

where $f^{\mathrm{n.p.}}_{\mathrm{conv}}$ denotes convolution without periodic boundary conditions. The teacher network has the same architecture, $\tau_{\mathrm{conv}}(\mathbf{x}) \coloneqq f^{\mathrm{n.p.}}_{\mathrm{conv}}(\boldsymbol{\theta}_0, \mathbf{x})$, and the Hessian is evaluated at the global minimum, $\mathsf{H}_{\mathrm{conv}} \coloneqq \partial^2_{\boldsymbol{\theta}}\mathcal{L}_{\mathrm{conv}}\big|_{\boldsymbol{\theta}=\boldsymbol{\theta}_0}$. In this setting, the Hessian coincides with the Fisher matrix of the network. The data is drawn from a normal distribution, $\mathbf{x} \sim \mathcal{N}(0, \mathbb{1}_{N+K-1})$, reflecting the higher-dimensional input compared to the periodic case.

This dimensionality difference clarifies why the symmetries are no longer exact for nonperiodic boundaries. Although the network remains linear in the input, it parametrizes a linear map from $N + K - 1$ dimensions to one dimension using $N + K$ parameters. In the periodic case, the map was from $N$ dimensions to one, implying at least $K$ redundant parameters, whereas in the nonperiodic case there is only a single redundant parameter. This corresponds to the one exact symmetry that remains: the trivial rescaling between first and second layer.

Nevertheless, numerical investigations of the Hessian for a randomly initialized teacher in the nonperiodic student–teacher setup reveal that the $K$ smallest eigenvalues are associated with eigenvectors having a high overlap with the symmetry generators of the corresponding periodic network (see Fig. S3). Thus, also in this convolutional network, small but finite Hessian eigenvalues are directly connected to approximate architectural symmetries.

---

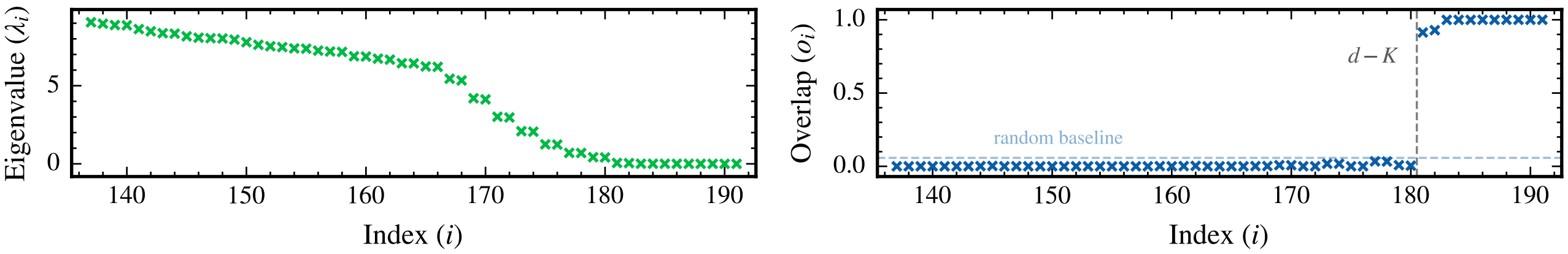


FIG. S3. Convolutional student–teacher network with input width 200, hidden width $N = 190$, and kernel width $K = 11$ employing mean-squared error loss. Results for a single draw of teacher weights from a Gaussian with variance 1/width. The Hessian is evaluated at identical student parameters. Left: 60 smallest Hessian eigenvalues of the loss Eq. (S31) of the nonperiodic network. Right: The overlap between eigenvectors of the nonperiodic network Hessian and the $K = 11$ symmetry directions of a corresponding periodic network is shown for the 60 smallest eigenvalues. Although boundary conditions break the exact symmetry, the smallest $K$ modes retain near-unity overlap, indicating weakly broken symmetries.


\fi

\end{document}